\documentclass[11pt]{article}

\usepackage[final]{acl}

\usepackage{times}
\usepackage{latexsym}

\usepackage[T1]{fontenc}
\usepackage[utf8]{inputenc}

\usepackage{microtype}

\usepackage{inconsolata}

\usepackage{graphicx}
\usepackage{amsmath}
\usepackage{amssymb}
\usepackage{booktabs}
\usepackage{multirow}
\usepackage{subcaption}

\usepackage{algorithm}
\usepackage{algorithmic}
\usepackage{listings}
\usepackage{longtable}
\usepackage{newfloat}
\usepackage{listings}
\DeclareCaptionStyle{ruled}{labelfont=normalfont,labelsep=colon,strut=off} % DO NOT CHANGE THIS
\floatstyle{ruled}
\newfloat{listing}{tb}{lst}{}
\floatname{listing}{Listing}

\usepackage{booktabs}
\usepackage{multirow}
\usepackage{tabularx}
\usepackage[most]{tcolorbox}
\usepackage{xcolor}

\usepackage{makecell}

\definecolor{gain}{RGB}{35,120,70}
\definecolor{drop}{RGB}{180,65,55}

\newcommand{\gain}[1]{%
  \textcolor{gain}{\scriptsize +#1}}
\newcommand{\dropv}[1]{%
  \textcolor{drop}{\scriptsize -#1}}
\newcommand{\same}{%
  \textcolor{gray}{\scriptsize --}}

\newcommand{\gcell}[2]{%
  \makecell[c]{#1 \\[-2pt] \gain{#2}}}
\newcommand{\dcell}[2]{%
  \makecell[c]{#1 \\[-2pt] \dropv{#2}}}
\newcommand{\scell}[1]{%
  \makecell[c]{#1 \\[-2pt] \same}}

\usepackage{CJKutf8}

\tcbset{
    promptbox/.style={
        enhanced,
        breakable,
        colback=white,
        colframe=black!60,
        coltitle=blue!70!black,
        boxrule=0.5pt,
        arc=2mm,
        left=2mm,
        right=2mm,
        top=1mm,
        bottom=1mm,
        before skip=6pt,
        after skip=6pt,
        listing only,
        listing options={
            basicstyle=\ttfamily\small,
            breaklines=true,
            breakatwhitespace=true,
            columns=fullflexible,
            keepspaces=true,
            showstringspaces=false
        }
    }
}
\title{AudiPsy: Evidence-Grounded Multimodal Reasoning over Spoken Counseling Conversations for Depression Assessment}

\author{
 \textbf{Chuanbo Zhu\textsuperscript{1}},
 \textbf{Run Luo\textsuperscript{3}},
 \textbf{Weibo Zhao\textsuperscript{4}},
 \textbf{Yue Li\textsuperscript{2}},
\\
 \textbf{Kun Qian\textsuperscript{5}},
 \textbf{Qiang Qu\textsuperscript{2}},
 \textbf{Yike Guo\textsuperscript{1}},
\textbf{Minghuan Tan\textsuperscript{2}}\NoHyper\thanks{Corresponding authors}\endNoHyper,
\textbf{Wei Xue\textsuperscript{1}}\footnotemark[\value{footnote}]
\vspace{0.2em}
\\
 \textsuperscript{1} The Hong Kong University of Science and Technology, \\
 \textsuperscript{2} Shenzhen Institutes of Advanced Technology, Chinese Academy of Sciences, \\
 \textsuperscript{3} National University of Singapore,
 \textsuperscript{4} University of the Chinese Academy of Sciences, \\
 \textsuperscript{5} Beijing Institute of Technology
\\
\small{
  \textbf{Correspondence:}
  \href{mailto:czhuat@connect.ust.hk}{czhuat@connect.ust.hk},
  \href{mailto:mh.tan@siat.ac.cn}{mh.tan@siat.ac.cn},
  \href{mailto:weixue@ust.hk}{weixue@ust.hk}
}
}

\begin{document}
\begin{CJK*}{UTF8}{gbsn}

\maketitle
\begin{abstract}
Depression assessment from counseling speech requires integrating textual and acoustic psychological evidence across turns, yet existing end-to-end classification approaches lack explicit multimodal evidence interpretation and context-aware reasoning.
We propose AudiPsy, an evidence-grounded omni-LLM for depression assessment over spoken counseling conversations.
AudiPsy employs an Observation--Assessment--Summary (OAS) paradigm to integrate dialogue transcripts, textual and acoustic emotions, and textual–acoustic emotion incongruence into structured reasoning. OAS incrementally organizes multimodal observations, performs contextualized assessment, and maintains an evolving dialogue summary for subsequent turns. To support this framework, we construct AudiPsy-Counsel, a synthetic spoken counseling conversation corpus with multimodal context augmentation, and develop a unified three-stage curriculum for depression-oriented reasoning.
Experiments on DAIC-WOZ demonstrate superior depression assessment performance over specialized depression detection methods and general-purpose speech/omni-LLMs, along with improved reasoning quality, while supplementary evaluation on EATD-Corpus supports AudiPsy's applicability to naturalistic conversations.
\end{abstract}
\section{Introduction}

Depression affects more than 322 million people worldwide and remains one of the leading causes of disability~\cite{who2025depression}.
Depression assessment is typically conducted through counseling conversations, where human counselors observe seekers' behaviors, interpret cues, accumulate evidence and reach a judgment~\cite{mao2023systematic}. Recent advances in AI for medical diagnosis have shifted from prediction toward evidence-based reasoning~\cite{wang2025medical,zhu2026diagnosisarena}. 
In the domain of mental health, automated depression assessment similarly calls for evidence-grounded multimodal reasoning across evolving spoken conversations.

Existing depression detection systems predominantly formulate the assessment as a multimodal classification problem, directly mapping linguistic and acoustic representations to diagnostic labels~\cite{wu2023self,zhang2024llms,zhang2025speecht}. Although effective on benchmark datasets, this paradigm suffers from two limitations. 
First, at the \emph{representation level}, most multimodal methods rely on low-level acoustic features or latent representations whose psychological meaning remains largely implicit.
Second, at the \emph{process level}, existing approaches typically aggregate multimodal representations into a holistic prediction without interpreting them under the conversational context.
These two limitations are coupled: the model cannot state \emph{what} it observes if it relies on implicit representations; the model cannot explain \emph{how} the observed evidence supports its assessment without an interpretive process. We therefore address them jointly.

\begin{figure*}[t!]
    \centering
    \includegraphics[width=\textwidth]{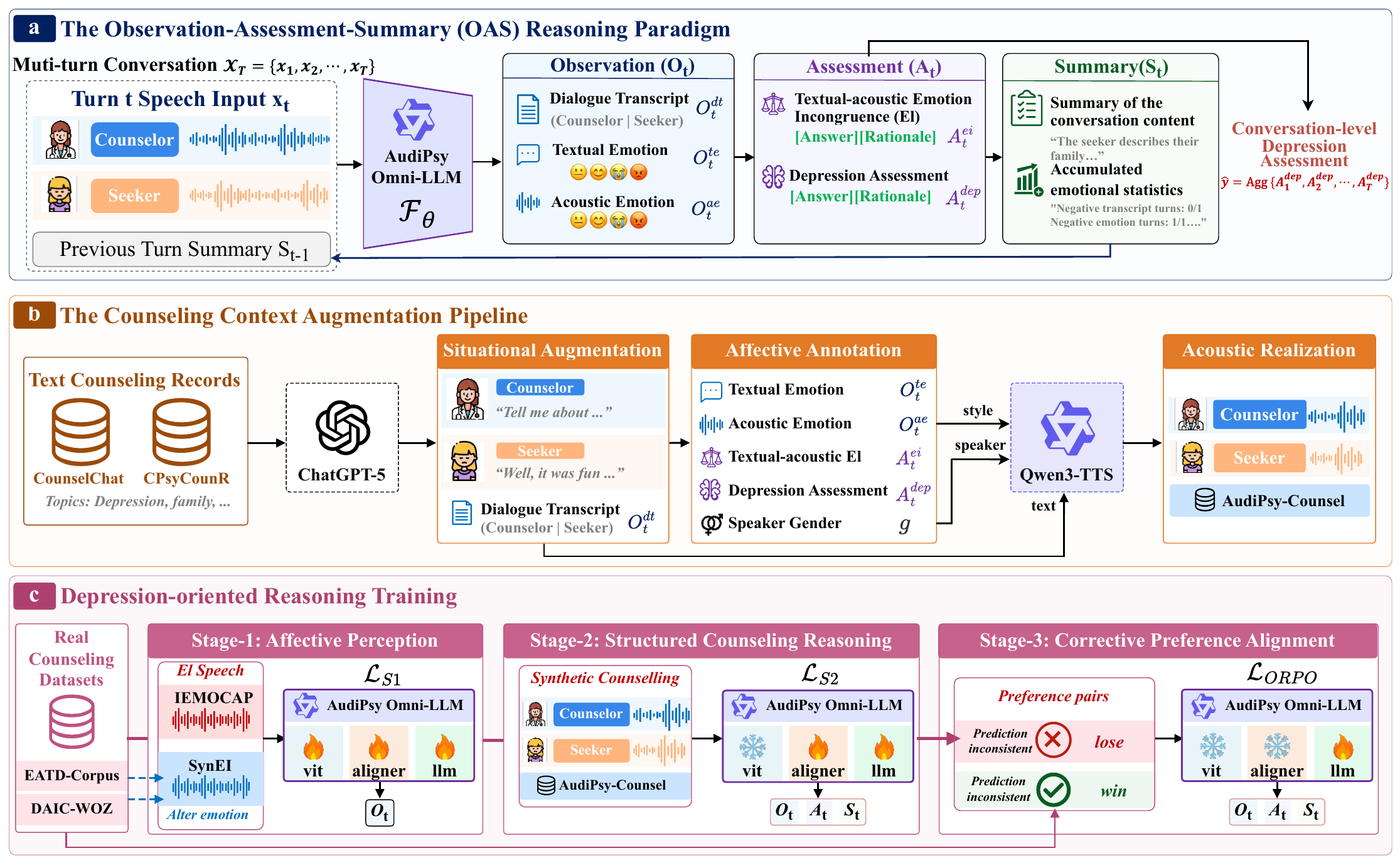}
    \caption{Overview of the AudiPsy, from structured multimodal reasoning to data augmentation and model training.}
    \label{fig:main}
\end{figure*}

For \emph{what} the model reasons over, we ground the assessment in affective evidence: textual emotion, acoustic emotion, and the incongruence between the two, expressed as affective concepts in natural language so that an LLM can reason over them.
Psychological studies have shown that individuals with depression often exhibit atypical emotional patterns~\cite{clark1991tripartite}, and inconsistency between semantic and acoustic emotion, as described by the emotion context insensitivity (ECI) theory~\cite{bylsma2008meta,wu2024depression}.
Recent studies have observed that modeling this emotional incongruence improves depression detection performance~\cite{su2026investigating}.
This evidence is psychologically interpretable and naturally amenable to language-model reasoning.

For \emph{how} the evidence is interpreted, following the structure of clinical progress notes~\cite{Weed1968medical}, we introduce the Observation–Assessment–Summary (OAS) paradigm for contextual reasoning over multi-turn spoken counseling conversations. At each conversational turn, OAS extracts structured semantic and acoustic observations (O), interprets them in light of accumulated dialogue context and textual–acoustic emotion incongruence (A), and updates a compact dialogue summary (S) to guide subsequent assessment.
This incremental process preserves the contextual meaning of psychological evidence while reducing the burden of long-context reasoning for current omni-LLMs~\cite{chen2026audio,wang2026humdial}.

To this end, we propose \textbf{AudiPsy}, an omni-LLM for depression assessment, which explicitly reasons over spoken counseling conversations in terms of multimodal affective evidence based on the proposed \textbf{Observation--Assessment--Summary (OAS)} reasoning paradigm.
To overcome the scarcity of real-world counseling speech corpus, we further develop a counseling context augmentation pipeline and a unified three-stage curriculum for depression-oriented reasoning.
Experiments on DAIC-WOZ~\cite{diac-woz}, featuring structured spoken counseling conversations, demonstrate superior depression assessment and reasoning performance. Additional evaluation on EATD-Corpus~\cite{eatd}, comprising naturalistic conversations, further examines AudiPsy's applicability across conversational settings.
Our contributions are summarized as follows:

\begin{itemize}
    \item We propose \textbf{AudiPsy}, an evidence-grounded omni-LLM for depression assessment through structured \textbf{OAS} reasoning.

    \item We construct \textbf{AudiPsy-Counsel}, a bilingual synthetic counseling speech corpus with OAS annotations and controlled emotion-incongruence cases.

    \item We develop a unified three-stage curriculum for \textbf{Depression-oriented Reasoning}, integrating affective perception, structured reasoning, and preference alignment.

    \item Experiments demonstrate improved depression assessment and reasoning quality, while analyses reveal how AudiPsy perceive, utilize, and interpret affective evidence.
\end{itemize}

\section{Related Work}
Automated depression assessment over spoken counseling conversations leverages linguistic, acoustic, and multimodal signals to identify mental health markers. Existing methods can be broadly categorized into three areas:
(1) Text-based approaches utilize transcripts to capture depressive states. Early methods rely on affective lexicons or neural representations~\cite{mallol2019hierarchical,xezonaki2020affective}, while recent frameworks employ large language models (LLMs) through prompting and instruction tuning~\cite{xu2024mental,liu2025enhanced}.
(2) Speech-based approaches identify depression-related cues from acoustic characteristics such as prosody and spectral features~\cite{sardari2022audio, wu2023self}, with recent methods incorporating data augmentation and retrieval-based reasoning to improve robustness~\cite{liang2025depressgen,zhang2025speecht}.
(3) Multimodal approaches integrate linguistic and acoustic signals through feature fusion~\cite{eatd,chen2024depression} or LLM-based modeling of textual and speech-derived representations~\cite{zhang2024llms,zhang2025speecht}.
Despite this progress, existing methods predominantly formulate depression assessment as end-to-end classification, leaving the psychological meaning of multimodal representations largely implicit and lacking contextualized interpretation of evidence as the conversation unfolds.

\section{Methodology}

As illustrated in Figure~\ref{fig:main}, AudiPsy combines the OAS reasoning paradigm with counseling context augmentation and a three-stage training curriculum for depression-oriented reasoning. Together, these components enable interpretable, context-aware multimodal reasoning for depression assessment.

\subsection{OAS Structured Reasoning Paradigm}
\label{sec:oas}

We propose an \textbf{Observation--Assessment--Summary (OAS)} reasoning paradigm that decomposes multimodal depression reasoning into three structured fields, as illustrated in Figure~\ref{fig:main}a.
Given a counseling speech with $T$ dialogue turns, we denote it as $\mathcal{X}=\{x_1,x_2,\ldots,x_T\}$, where each turn is represented by $x_t=(c_t,s_t)$, with $c_t$ and $s_t$ denoting the counselor's and seeker's spoken utterances, respectively. The ordering is used only for notation and does not imply which speaker initiates the dialogue.
At each turn, AudiPsy maps the current dialogue turn and the previous summary state to an observation, an assessment, and an updated summary:
\begin{equation}
(O_t,A_t,S_t)
=
\mathcal{F}_{\theta}(x_t,S_{t-1}),
\end{equation}
where $\mathcal{F}_{\theta}$ denotes AudiPsy parameterized by $\theta$, and $S_{t-1}$ summarizes the accumulated dialogue context from previous turns. The three fields are defined as follows.

The \textbf{Observation} $O_t=(O_t^{\mathrm{dt}},O_t^{\mathrm{te}},O_t^{\mathrm{ae}})$ consists of the dialogue transcript $O_t^{\mathrm{dt}}$, the seeker's textual emotion $O_t^{\mathrm{te}}$ and acoustic emotion $O_t^{\mathrm{ae}}$. Together, they provide explicit and interpretable multimodal evidence for subsequent reasoning.

The \textbf{Assessment} $A_t=(A_t^{\mathrm{ei}},A_t^{\mathrm{dep}})$ contains the textual--acoustic emotion incongruence (EI) assessment $A_t^{\mathrm{ei}}$ and the depression assessment $A_t^{\mathrm{dep}}$, each consisting of a binary prediction answer together with its supporting rationale.
Therefore, the depression assessment can be formulated as a conditional dependency
$p_{\theta}(A_t^{\mathrm{dep}}\mid A_t^{\mathrm{ei}},O_t,S_{t-1})$, indicating that the reasoning $A_t^{\mathrm{dep}}$ is grounded by evidence $A_t^{\mathrm{ei}},O_t$ and $S_{t-1}$.

The \textbf{Summary} $S_t$ recursively maintains a compact dialogue state, including a concise summary of the counseling content together with accumulated emotional statistics and affective patterns.

Finally, the dialogue-level depression prediction is obtained by aggregating the turn-level depression assessments:
\begin{equation}
\hat{y}
=
\mathrm{Agg}
\left(
A_1^{\mathrm{dep}},A_2^{\mathrm{dep}},\ldots,A_T^{\mathrm{dep}}
\right),
\end{equation}
where $\mathrm{Agg}(\cdot)$ denotes a consistency-based aggregation function.
In our implementation, $\mathrm{Agg}(\cdot)$ uses majority voting over turn-level binary predictions for robust conversation-level aggregation.
Detailed output templates and field definitions are provided in Appendix~\ref{appendix: oas}.

\subsection{Counseling Context Augmentation}

Real-world speech-based counseling data is extremely scarce due to privacy concerns and acquisition costs, while text-only counseling resources are considerably abundant, but lack the affective context and its acoustic realization required for multimodal depression reasoning.
To bridge this gap, we formulate data construction as a \emph{context augmentation} problem. Starting from publicly available text-only counseling records, we progressively enrich them along three orthogonal dimensions: \emph{situational, affective, and acoustic}, so that each resulting sample supplies the contextual evidence consumed by the OAS paradigm, as shown in Figure~\ref{fig:main}b.
Although synthetic, the resulting dataset, AudiPsy-Counsel, provides structured supervision for learning how to perceive and interpret multimodal evidence to support depression assessment across dialogue turns.

\begin{figure}[t]
    \centering
    \includegraphics[width=\linewidth]{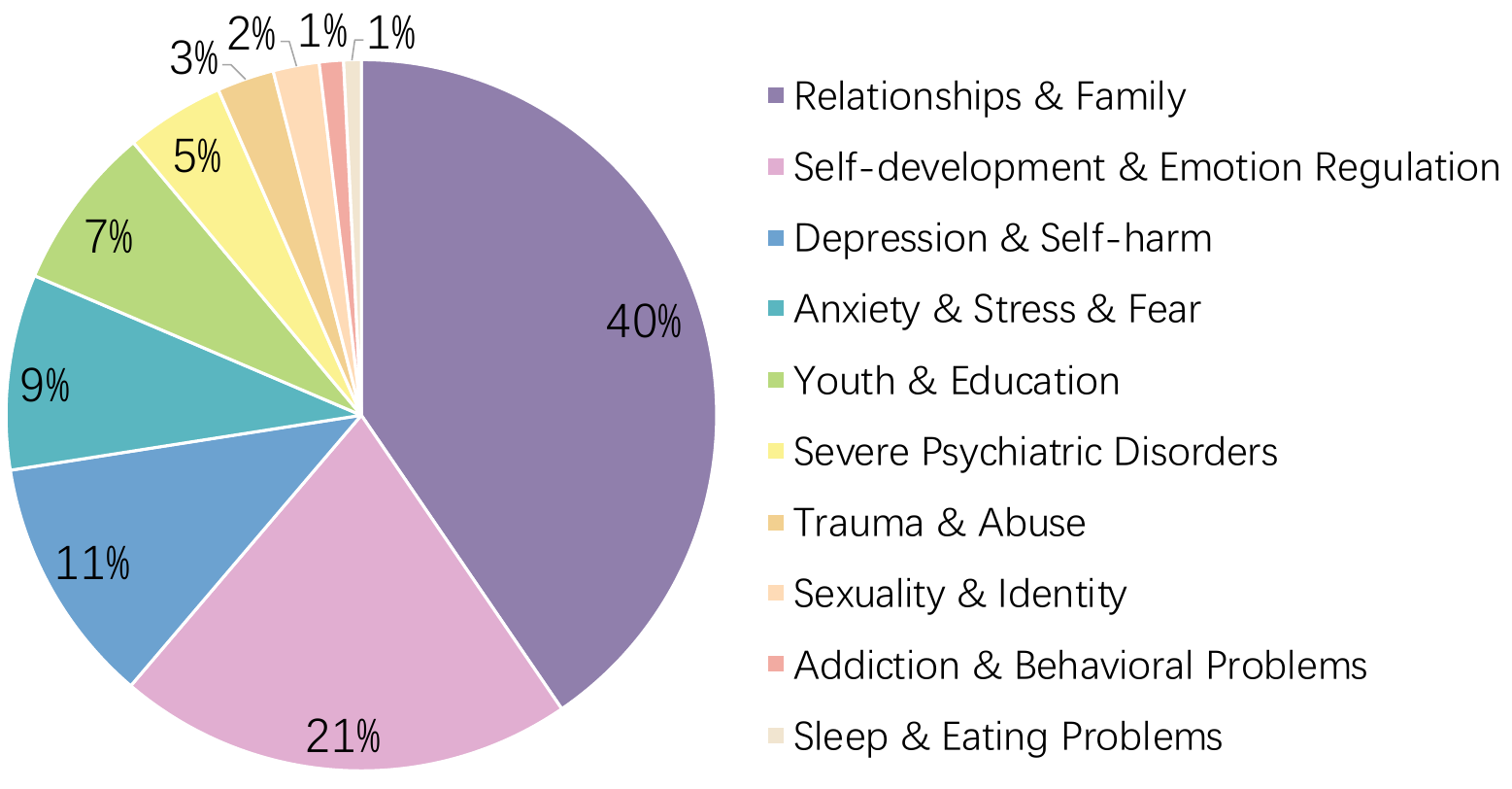}
    \caption{Topic distribution of AudiPsy-Counsel.}
    \label{fig:topics}
\end{figure}

\paragraph{Situational augmentation.}
The seed corpus is an union of two public text counseling corpora,
CounselChat~\cite{bertagnolli2020counsel} and CPsyCounR~\cite{zhang-etal-2024-cpsycoun}.
We employ GPT-5~\cite{openai_gpt5_2025} as a situational re-writer which maps each counseling record into a bilingual dialogue transcript $O^{\mathrm{dt}}$
between a counselor and a seeker.
The rewriting preserves the original clinical issue while diversifying emotional conditions and interaction trajectories, resulting in richer situational variations while preserving the natural distribution of counseling topics, including depression, relationships, self-development, and more (Figure~\ref{fig:topics}).

\paragraph{Affective annotation.}
For each utterance, GPT-5 also acts as an affective annotator to produce textual and acoustic emotion annotations $O_t^{\mathrm{te}}$ and $O_t^{\mathrm{ae}}$, together with the textual--acoustic emotion incongruence annotation $A_t^{\mathrm{ei}}$, while the seeker gender $g$ is annotated at the dialogue level.
Together, these annotations provide explicit ground-truth multimodal evidence including textual--acoustic emotion incongruence, which is rarely available and difficult to verify in organic data.

Conditioned on the annotated $O_t$, $A^{ei}_t$, and $S_{t-1}$, together with the original counseling record, the LLM generates the depression assessment $A^{dep}_t$ following a reasoning skill informed by the ECI theory and PHQ-8 diagnostic algorithm~\cite{kroenke2009phq}. Since each assessment is conditioned on explicit observations and guided by these principles, the resulting reasoning is grounded in multimodal evidence. Consequently, each dialogue turn is annotated under the complete OAS paradigm.

\paragraph{Acoustic realization.}

The annotated scripts are rendered into emotional speech with the
prompt-controlled text-to-speech synthesis model Qwen3-TTS~\cite{Qwen3-TTS},
where dialogue transcripts $O^{dt}_t$ serve as synthesis inputs, the acoustic
emotion descriptions $O^{ae}_t$ serve as speech-style prompts, and the speaker gender $g$ serves as the speaker selection reference.

Consequently, the pipeline yields AudiPsy-Counsel, denoted as $\mathcal{D}_{\mathrm{apc}}$, a synthetic bilingual speech-counseling dataset with 2,400 conversations for downstream model training.
All the details of the dataset construction, including prompt templates, quality-control procedures, and example instances are provided in Appendix~\ref{appendix:audipsy-counsel}.

\subsection{Depression-oriented Reasoning}
\label{subsec: depression reasoning}

To equip $\mathcal{F}_{\theta}$ with the capability of evidence-grounded multimodal reasoning for depression assessment, we optimize it through a three-stage training curriculum, as illustrated in Figure~\ref{fig:main}c. The curriculum progressively develops affective perception, structured counseling reasoning, and corrective preference alignment.

\paragraph{Stage-1: Affective perception.}
We first train $\mathcal{F}_{\theta}$ to learn the observation field $O_t$ from emotional speech to enable robust multimodal affective perception under both emotion-congruent and emotion-incongruent conditions.
Specifically, we use IEMOCAP~\cite{busso2008iemocap} as the spontaneous corpus, where we annotate the textual polarity of each utterance and construct a balanced training set by randomly downsampling each textual--acoustic emotion label combination. 

We further construct a Synthetic Emotion-Incongruence (SynEI) dataset, a depression-related emotional speech corpus derived from the training split of DAIC-WOZ and EATD-Corpus, by altering the speech emotion through speech re-synthesis~\cite{zhou2026indextts2} while preserving the original text and speaker characteristics.
Construction and dataset details are provided in Appendices~\ref{appendix:iemocap} and~\ref{appendix:synEI}.
Accordingly, the Stage-1 objective is defined as:
\begin{equation}
\mathcal{L}_{\mathrm{S1}}
=
-
\mathbb E_{\mathcal D_{\mathrm{emo}}}
\log p_\theta(O_t\mid x_t).
\end{equation}

\paragraph{Stage-2: Structured counseling reasoning.}
We then continue supervised fine-tuning on AudiPsy-Counsel $D_{apc}$ to adapt the model to depression-oriented counseling reasoning. 
$\mathcal{F}_{\theta}$ learns the complete OAS reasoning process with the Stage-2 objective defined as:
\begin{equation}
    \mathcal{L}_{\mathrm{S2}}
    = -\,\mathbb{E}_{\mathcal{D}_{\mathrm{apc}}}
    \log p_{\theta}
    \left( O_t, A_t, S_t \mid x_t, S_{t-1} \right).
\end{equation}

% \paragraph{Stage-3: Real counseling alignment.}
\paragraph{Stage-3: Corrective preference alignment.}
Finally, we align $\mathcal{F}_{\theta}$ on real depression-assessment data (DAIC-WOZ; EATD-Corpus), denoted as $\mathcal{D}_{\mathrm{real}}$. Considering its limited scale, we adopt the lightweight ORPO objective~\cite{hong2024orpo}, which jointly performs supervised learning and preference optimization in a single stage.
For EATD-Corpus, the counselor speech signals $c_t$ are reconstructed using Qwen-TTS, as they are unavailable in the original dataset.
During ORPO training, the Stage-2 output is treated as the rejected response, while a preferred response is regenerated using a pretrained speech emotion recognition (SER) model~\cite{speechvad-model} together with the AudiPsy-Counsel affective annotation pipeline conditioned on the ground-truth depression label.
The model is optimized using the standard ORPO objective:
\begin{equation}
\mathcal{L}_{\mathrm{ORPO}}
=
\mathbb{E}_{\mathcal{D}_{\mathrm{real}}}
\left[
\mathcal{L}_{\mathrm{SFT}}
+
\lambda\mathcal{L}_{\mathrm{OR}}
\right].
\end{equation}

\section{Experiments}

\begin{table*}[t]
\centering
\small
\setlength{\tabcolsep}{5pt}

\begin{tabular}{lccccc}
\toprule
\textbf{Method}
& \textbf{MCC} & \textbf{F1}
& \textbf{Recall} & \textbf{Precision}
& \textbf{Specificity} \\
\midrule

BiLSTM + GRU$^{*}$
& 0.11 & 0.53 & 0.56 & 0.56 & -\\

WavLM Layer 10~\cite{wu2023self}
& - & 0.72 & - & - & -\\

Llama2 + Acoustic Landmarks~\cite{zhang2024llms}
& - & 0.72 & - & - & -\\

SpeechT-RAG~\cite{zhang2025speecht}
& - & 0.69 & - & - & -\\

\midrule

Kimi-Audio-7B-Instruct~\cite{kimiteam2025kimiaudiotechnicalreport}
& 0.00 & 0.00 & 0.00 & 0.00 & 0.00\\

MOSS-Audio-8B-Instruct~\cite{yang2026mossaudiotechnicalreport}
& 0.36 & 0.62 & 0.75 & 0.53 & 0.62\\

MOSS-Audio-8B-Thinking~\cite{yang2026mossaudiotechnicalreport}
& 0.52 & 0.71 & 0.92 & 0.58 & 0.62\\

\midrule

Gemini 2.5 Flash~\cite{comanici2025gemini}
& 0.61 & 0.76 & 0.92 & 0.65 & 0.71\\

Gemini 2.5 Pro~\cite{comanici2025gemini}
& 0.65 & 0.79 & 0.92 & 0.69 & 0.76\\

Qwen2.5-Omni-7B~\cite{Qwen2.5-Omni}
& 0.24 & 0.42 & 0.33 & 0.57 & 0.82\\

Qwen3-Omni-30B-A3B-Instruct~\cite{xu2025qwen3}
& 0.19 & 0.56 & \textbf{1.00} & 0.39 & 0.10\\

Qwen3-Omni-30B-A3B-Thinking~\cite{xu2025qwen3}
& 0.00 & 0.00 & 0.00 & 0.00 & \textbf{1.00}\\

\midrule

Gemini 2.5 Pro (OAS)
& 0.00 & 0.53 & \textbf{1.00} & 0.36 & 0.00\\

Qwen2.5-Omni-7B (OAS)
& 0.35 & 0.62 & 0.83 & 0.50 & 0.52\\

Qwen3-Omni-30B-A3B-Instruct (OAS)
& 0.24 & 0.57 & \textbf{1.00} & 0.40 & 0.14\\

\midrule

\textbf{AudiPsy (Ours)}
& \textbf{0.70} & \textbf{0.81} & 0.92 & \textbf{0.73} & 0.81\\

\phantom{AudiPsy} w/o $\mathcal{L}_{\mathrm{S1}}$
& 0.58 & 0.74 & 0.83 & 0.67 & 0.76\\

\phantom{AudiPsy} w/o $\mathcal{L}_{\mathrm{ORPO}}$
& 0.65 & 0.79 & 0.92 & 0.73 & 0.76\\

\phantom{AudiPsy} w/o $\mathcal{L}_{\mathrm{S1}} + \mathcal{L}_{\mathrm{ORPO}}$
& 0.61 & 0.75 & 1.00 & 0.60 & 0.62\\

\phantom{AudiPsy} w/o $\mathcal{L}_{\mathrm{S2}} + \mathcal{L}_{\mathrm{ORPO}}$
& 0.53 & 0.71 & 0.83 & 0.63 & 0.71\\

\phantom{AudiPsy} w/o Emotion Incongruence
& 0.58 & 0.73 & 0.92 & 0.81 & 0.76\\

\phantom{AudiPsy} w/ Unstructured Rationale
& 0.65 & 0.69 & 0.92 & 0.79 & 0.76\\

\bottomrule
\end{tabular}

\caption{
Depression assessment performance on the
DAIC-WOZ dataset in terms of MCC, F1, recall, precision,
and specificity.
}
\label{tab:main_results}
\end{table*}

% \begin{figure*}[tbp]
%     \centering
%     \includegraphics[width=\textwidth]{figures/Figure_1.png}
%     \caption{
%         Human evaluation results on the DAIC-WOZ dataset.
%         (a) Relative scores of each method across four evaluation dimensions
%         (EG, CP, MMI, and CEI), calculated as deviations from the mean score
%         across methods.
%         (b) Win--Tie--Lose percentages of AudiPsy against four baseline methods
%         across the same evaluation dimensions.
%     }
%     \label{fig:human_evaluation}
% \end{figure*}

We evaluate AudiPsy from three perspectives: (1) depression assessment and human reasoning quality evaluation on DAIC-WOZ; (2) supplementary depression assessment on EATD-Corpus; and (3) multimodal evidence perception, utilization, and interpretation through speech emotion recognition, evidence utilization analysis, and reasoning case studies.

\paragraph{Training}
AudiPsy is initialized from Qwen2.5-Omni-7B~\cite{Qwen2.5-Omni} and trained with the proposed three-stage curriculum using the ms-swift framework~\cite{zhao2024swift} on 4 NVIDIA H800 GPUs. Stage-1 and -2 perform LoRA-based supervised fine-tuning on the affective perception corpus $\mathcal{D}_{\mathrm{emo}}$ (2,109 samples) and the AudiPsy-Counsel corpus $\mathcal{D}_{\mathrm{apc}}$ (2,400 samples), respectively. Stage 3 further aligns the model using ORPO on the depression-assessment corpus $\mathcal{D}_{\mathrm{real}}$ (280 conversations after conversation-level resampling to mitigate class imbalance).

\paragraph{Inference}
During inference, AudiPsy performs turn-by-turn reasoning, conditioning each turn on the previous summary. Dialogue-level predictions are obtained by majority voting, with ties resolved as \texttt{yes}. Greedy decoding (temperature = 0) ensures deterministic evaluation. Following prior work~\cite{wu2023self, zhang2024llms, zhang2025speecht}, we evaluate on the DAIC-WOZ development set (33 conversations: 12 depressed, 21 non-depressed) for direct comparability. Separately, we evaluate on the EATD-Corpus validation set (79 conversations: 11 depressed, 68 non-depressed) to assess performance across conversational settings. Further implementation details and prompts are provided in Appendix~\ref{appendix: training and inference}.

\subsection{Depression Assessment Performance}

\paragraph{Baselines}
We compare AudiPsy against representative speech-based binary depression classification methods, including BiLSTM+GRU~\cite{shen2022automatic} and WavLM Layer 10~\cite{wu2023self}, as well as multi-stage LLM pipelines, including Speech Landmarks+Llama2~\cite{zhang2024llms} and SpeechT-RAG~\cite{zhang2025speecht}. We reproduce BiLSTM+GRU using its official implementation and test set, while reporting published single-model F1 results for the other methods.

For speech LLMs, we select models of comparable scale to AudiPsy's Qwen2.5-Omni-7B backbone~\cite{kimiteam2025kimiaudiotechnicalreport,yang2026mossaudiotechnicalreport}. For omni-LLMs, we include the Gemini 2.5 series~\cite{comanici2025gemini} as representative proprietary models, together with Qwen2.5-Omni-7B~\cite{Qwen2.5-Omni} and the larger Qwen3-Omni-30B series~\cite{xu2025qwen3} from the same Omni family. A subset of these models is additionally evaluated under the OAS paradigm without task-specific fine-tuning. All evaluated models are listed in Table~\ref{tab:main_results}.

\paragraph{Metrics}
We evaluate AudiPsy from two perspectives: depression assessment performance and reasoning quality.
For depression assessment, we report Matthews Correlation Coefficient (MCC), binary F1, recall, precision, and specificity. In addition to standard binary classification metrics, we include MCC and specificity to provide a balanced measure under class imbalance, accounting for performance on both depressed and non-depressed individuals~\cite{chicco2020advantages}.

For reasoning quality, four practicing psychological counselors evaluate 5 methods on 10 class-balanced conversations from DAIC-WOZ, examining 250 LLM outputs per evaluator. Reasoning quality is rated along four dimensions: Evidence Grounding (EG), Clinical Plausibility (CP), Multi-Modal Integration (MMI), and Contextual Evidence Integration (CEI). Evaluators assigned integer scores on a 1–5 scale. To accommodate dimension-specific rating baselines, we report mean-centered scores and win--tie--loss results across methods for each dimension.
Detailed evaluation protocols are provided in Appendix~\ref{appendix:reasoning_metrics}.

\paragraph{Main results}
As shown in Table~\ref{tab:main_results}, AudiPsy attains the highest MCC,
F1, and precision, outperforming specialized speech-based
methods and general-purpose speech and omni-LLMs on DAIC-WOZ.
Notably, Kimi-Audio fails to generate outputs for long audio inputs, while Qwen3-Omni-30B-A3B-Thinking frequently declines to make depression predictions.

Simply applying OAS to off-the-shelf models does not consistently improve performance. Although OAS explicitly structures multimodal evidence, Qwen2.5-Omni-7B often produces inaccurate observations and assessments, while Gemini 2.5 Pro and Qwen3-Omni-Instruct exhibit high recall but limited MCC, suggesting a tendency toward positive depression predictions. 
These results reinforce the coupling between evidence representation and contextual interpretation: structured reasoning alone is insufficient without reliable affective perception and task-specific training.

\paragraph{Ablation study}
We conduct six ablation experiments to examine the contributions of
the training curriculum, emotion incongruence, and the structured OAS
representation. Removing individual training objectives or their
combinations generally degrades performance, showing the necessity of the full three-stage curriculum.
In particular, removing $\mathcal{L}_{\mathrm{S1}}$ leads to a clear
performance drop, highlighting the importance of affective perception
for subsequent depression reasoning.
Removing emotion incongruence also substantially degrades performance,
demonstrating its importance as a source of multimodal evidence.

We further analyze their effects on
affective perception and evidence utilization, respectively, in
Section~\ref{subsec:analysis}.
We also rewrite the structured OAS fields
into an unstructured natural-language rationale while preserving the
same evidence content and passing it across dialogue turns during
training. The resulting performance degradation shows that the benefit of OAS goes beyond rationale content
itself, with its explicit structure facilitating evidence organization
and information maintenance across dialogue turns.

\begin{figure}[t]
    \centering

    % (a) Relative Scores - Top
    \begin{subfigure}[t]{\columnwidth}
        \centering
        \includegraphics[width=\linewidth]{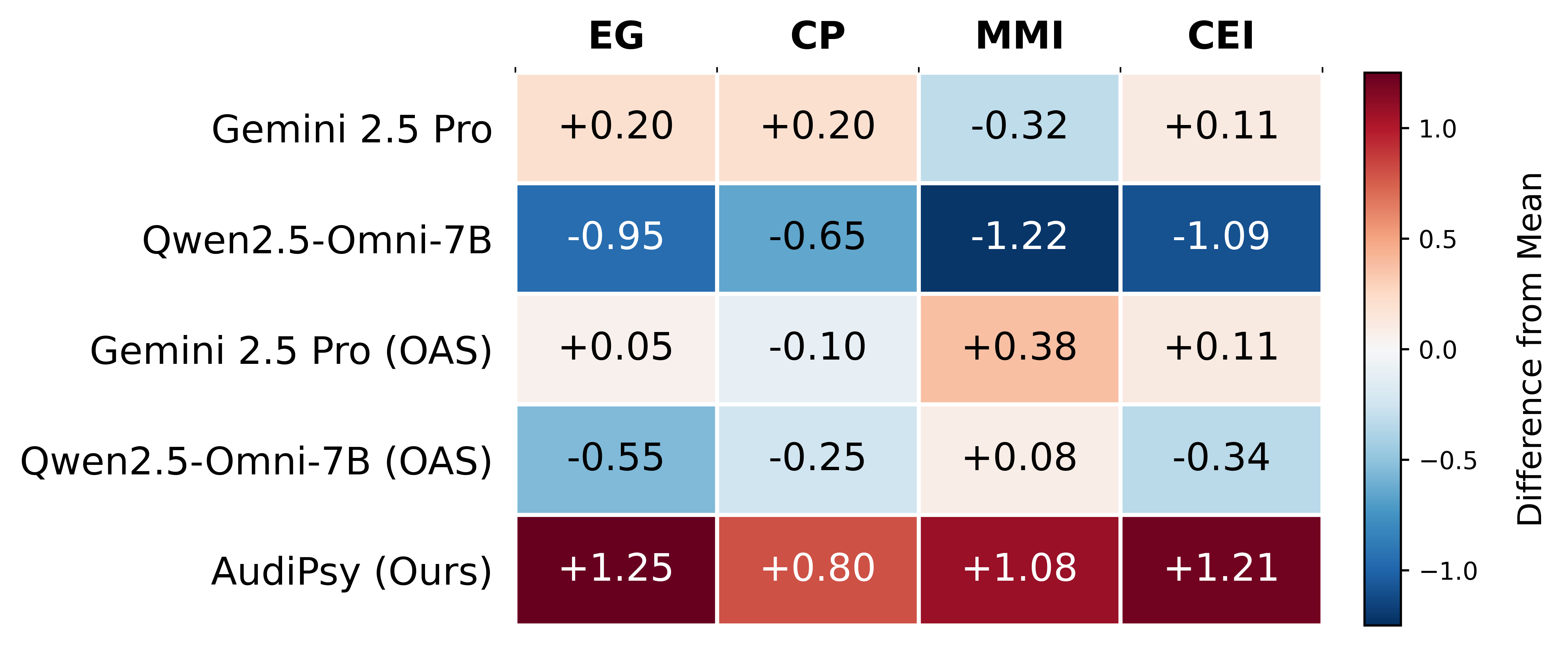}
        \caption{Relative scores.}
        \label{fig:relative_scores}
    \end{subfigure}

    \vspace{0.3cm}

    % (b) Win-Tie-Lose - Bottom
    \begin{subfigure}[t]{\columnwidth}
        \centering
        \includegraphics[width=\linewidth]{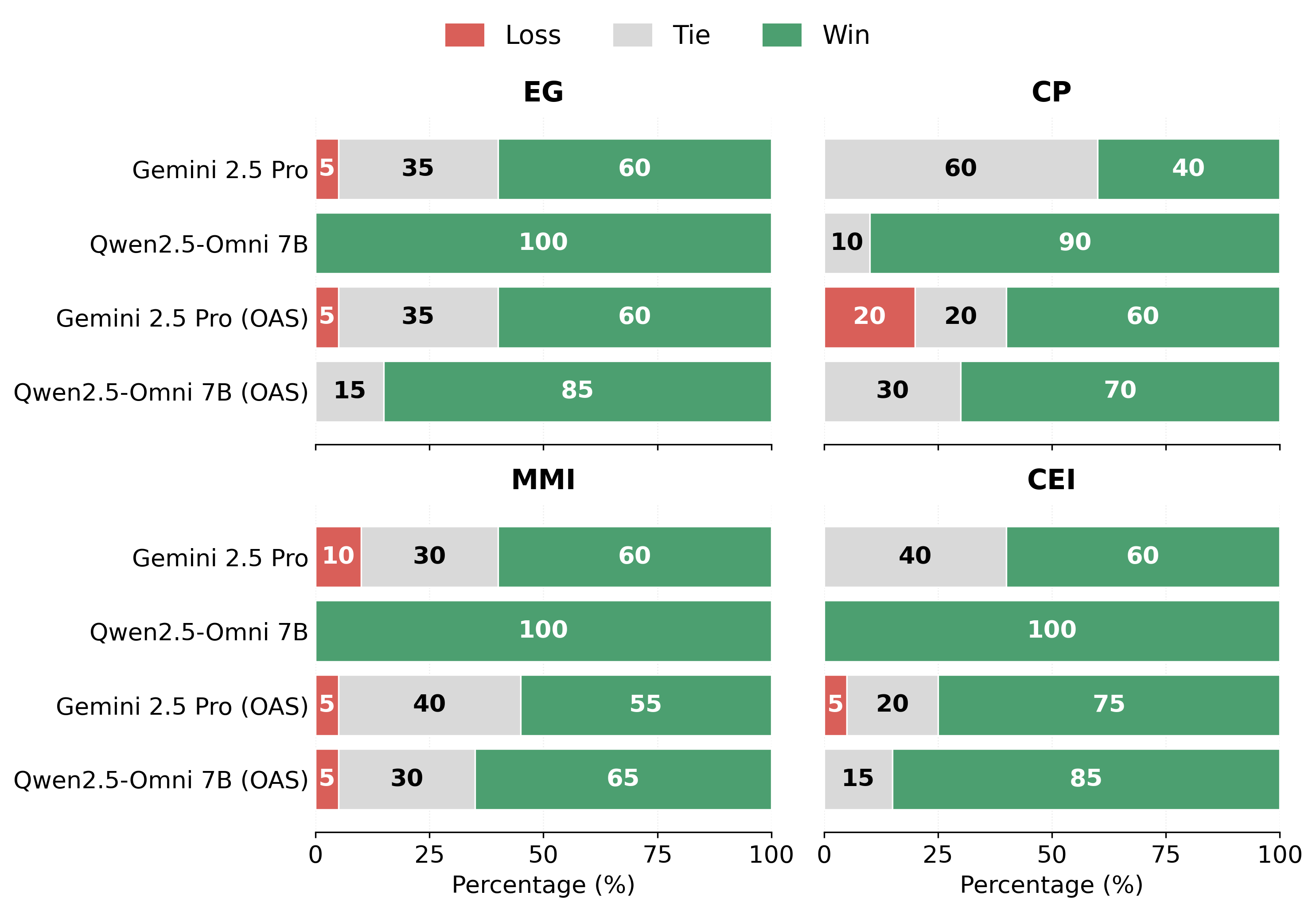}
        \caption{Win--Tie--Loss comparison.}
        \label{fig:win_tie_lose}
    \end{subfigure}

    \caption{
        Human evaluation results on the DAIC-WOZ dataset.
        (a) Relative scores of each method across four evaluation dimensions
        (EG, CP, MMI, and CEI), calculated as deviations from the mean score
        across methods.
        (b) Win--Tie--Loss percentages of AudiPsy against four baseline methods
        across the same evaluation dimensions.
    }
    \label{fig:human_evaluation}
\end{figure}

\subsection{Human Evaluation of Reasoning Quality}
\label{subsection: human eval}

As shown in Figure~\ref{fig:human_evaluation}, AudiPsy achieves the highest
reasoning quality across all evaluation dimensions, complementing its
strong performance on the binary depression assessment. For both Gemini 2.5 Pro and Qwen2.5-Omni-7B, applying OAS generally improves reasoning
quality, particularly in MMI and CEI, demonstrating the benefit of
explicitly organizing multimodal evidence and conversational context.
However, Gemini 2.5 Pro (OAS) shows a slight decrease in CP, suggesting
that structured reasoning alone does not necessarily improve the
interpretation of multimodal evidence for depression assessment.
Moreover, the win--tie--loss results show that AudiPsy is consistently
preferred over the baselines, indicating that its higher relative
scores reflect systematic improvements across evaluated cases rather
than gains driven by a small number of examples.

\begin{table}[t]
\centering
\small
\setlength{\tabcolsep}{5pt}
\begin{tabular}{lcc}
\toprule
\textbf{Method} & \textbf{MCC} & \textbf{F1} \\
\midrule
BiLSTM + GRU$^{*}$
& 0.19 & 0.32 \\
MOSS-Audio-8B-Thinking
& 0.27 & 0.37 \\
Gemini 2.5 Flash
& 0.13 & 0.29 \\

Gemini 2.5 Pro
& 0.23 & 0.33 \\

Qwen2.5-Omni-7B
& 0.11 & 0.21 \\

Qwen3-Omni-30B-A3B-Instruct
& 0.20 & 0.33 \\

Qwen3-Omni-30B-A3B-Thinking
& 0.32 & \textbf{0.40} \\
\midrule
\textbf{AudiPsy (Ours)}
& \textbf{0.41} & \textbf{0.40} \\

\quad w/o $\mathcal{L}_{\mathrm{S1}}$
& 0.28 & 0.17 \\

\quad w/o Emotion Incongruence
& -0.05 & 0.00 \\
\bottomrule
\end{tabular}
\caption{
Depression assessment
performance on EATD-Corpus in terms of MCC and F1.
}
\label{tab:eatd_results}
\end{table}

\subsection{Depression Assessment in Everyday Conversations}
To complement the primary DAIC-WOZ evaluation, we assess AudiPsy on EATD-Corpus, which contains naturalistic conversations with less structured diagnostic context, posing additional challenges for depression assessment, as corroborated by practicing psychological counselors.
As shown in Table~\ref{tab:eatd_results}, AudiPsy achieves the highest MCC and competitive F1, supporting its applicability across conversational settings.
Consistent with DAIC-WOZ, ablations highlight the importance of affective perception and emotion incongruence modeling, suggesting that modeling subtle affective cues and cross-modal relationships contributes to depression assessment in this setting.

% ---------- SER Table ----------
\begin{table*}[t]
\centering
\footnotesize
\setlength{\tabcolsep}{3.5pt}

\begin{tabular}{@{}lcccccccccccc@{}}
\toprule
\multirow{2}{*}{\textbf{Model}} &
\multicolumn{4}{c}{\textbf{Positive}} &
\multicolumn{4}{c}{\textbf{Neutral}} &
\multicolumn{4}{c}{\textbf{Negative}} \\
\cmidrule(lr){2-5}
\cmidrule(lr){6-9}
\cmidrule(lr){10-13}

& \textbf{Happy} & \textbf{Angry} & \textbf{Neutral} & \textbf{Sad}
& \textbf{Happy} & \textbf{Angry} & \textbf{Neutral} & \textbf{Sad}
& \textbf{Happy} & \textbf{Angry} & \textbf{Neutral} & \textbf{Sad} \\

\midrule

Gemini 2.5 Pro &
\textbf{0.58} & \textbf{0.40} & \textbf{0.64} & 0.62 &
0.47 & \textbf{0.75} & 0.61 & 0.43 &
0.57 & 0.83 & 0.47 & 0.80 \\

Qwen2.5-Omni-7B &
0.45 & 0.18 & 0.64 & 0.33 &
0.00 & 0.33 & 0.45 & 0.17 &
0.00 & 0.80 & 0.36 & 0.56 \\

\midrule

\textbf{AudiPsy (Ours)} &
\dcell{0.39}{.06} &
\scell{0.18} &
\dcell{0.44}{.20} &
\gcell{\textbf{0.67}}{.34} &
\gcell{\textbf{0.64}}{.64} &
\scell{0.33} &
\gcell{\textbf{0.67}}{.22} &
\gcell{\textbf{0.64}}{.47} &
\gcell{\textbf{0.63}}{.63} &
\gcell{\textbf{0.89}}{.09} &
\gcell{\textbf{0.67}}{.31} &
\gcell{\textbf{0.84}}{.28} \\

\addlinespace[2pt]

\phantom{\textbf{AudiPsy}} w/o $\mathcal{L}_{\mathrm{S1}}$ &
\dcell{0.44}{.01} &
\dcell{0.00}{.18} &
\dcell{0.53}{.11} &
\dcell{0.00}{.33} &
\gcell{0.56}{.56} &
\scell{0.33} &
\gcell{0.56}{.11} &
\gcell{0.29}{.12} &
\gcell{0.18}{.18} &
\dcell{0.67}{.13} &
\dcell{0.32}{.04} &
\gcell{0.62}{.06} \\

\bottomrule
\end{tabular}

\caption{
Per-class F1 scores across transcript polarity groups.
Smaller values below AudiPsy and its ablation indicate absolute F1
changes relative to the Qwen2.5-Omni-7B backbone;
\textcolor{gain}{green} and \textcolor{drop}{red} denote improvements
and degradations, respectively.
}
\label{tab:ser}
\end{table*}

\subsection{Probing the Assessment Behavior}
\label{subsec:analysis}

\paragraph{Perceptual foundation: cross-modal speech emotion recognition}
We first probe whether AudiPsy perceives speech emotion independently of textual sentiment, which is the perceptual foundation of incongruence analysis and affective evidence integration.
Table~\ref{tab:ser} reports
per-class F1 scores for four speech emotion categories on the IEMOCAP
test set, stratified by transcript polarity.
The Qwen2.5-Omni-7B backbone relies heavily on textual
affective cues, struggling when speech emotion conflicts with transcript polarity. This tendency persists without $\mathcal{L}_{\mathrm{S1}}$, whereas full AudiPsy improves emotion recognition, particularly for Neutral and Negative speech, consistent with the downstream ablation results in Table~\ref{tab:main_results}.
Meanwhile, Gemini 2.5 Pro achieves comparable emotion recognition but substantially weaker depression assessment following OAS paradigm, suggesting that accurate affective perception alone is insufficient without appropriate evidence interpretation.

\begin{figure}[]
    \centering
    \includegraphics[width=\linewidth]{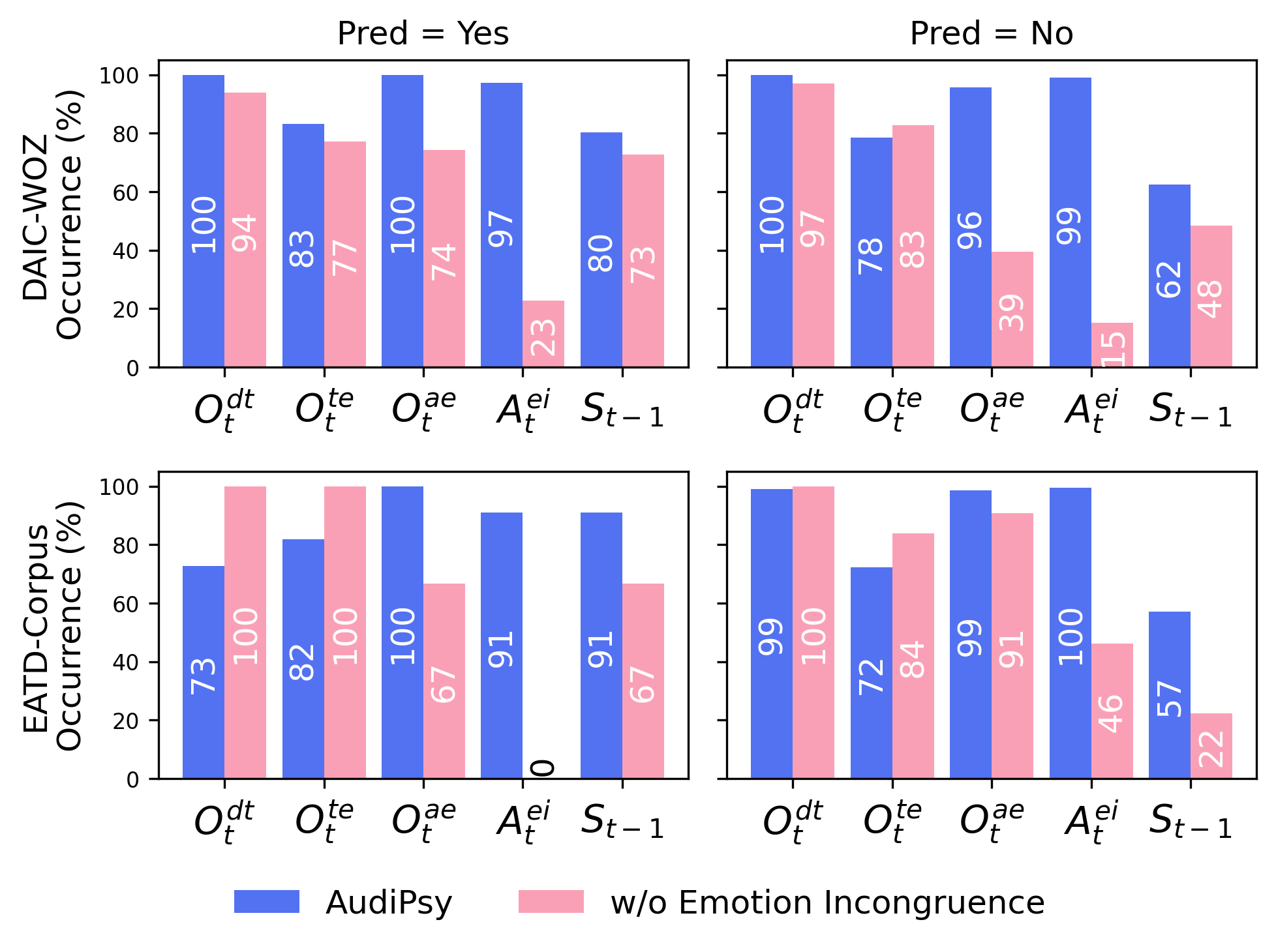}
    \caption{Occurrence rates of evidence types employed by AudiPsy and its ablation without Emotion Incongruence predicting ``Yes (Depressed)'' or ``No (Non-depressed)''.}
    \label{fig:evidence_rate}
\end{figure}

\paragraph{Evidence utilization: impact of emotion incongruence}

We next probe how explicit emotion incongruence modeling shapes the evidence used in assessment reasoning. Figure~\ref{fig:evidence_rate} summarizes the percentage of evidence types used in depression assessment reasoning across the DAIC-WOZ and EATD-Corpus test sets. We use GPT-5 to extract and categorize evidence types utilized in the rationale.
The full AudiPsy exploits all five types of evidence in a balanced manner, whereas removing explicit emotion incongruence analysis causes the model to rely more heavily on textual cues while using substantially fewer acoustic emotion cues and accumulated multimodal patterns.
This effect is particularly evident of the ``Yes'' prediction on EATD-Corpus, where less informative textual content leads the full model to naturally shift toward acoustic evidence, while the ablated model fails to exploit such complementary multimodal information.
Consistent with this shift in evidence utilization, removing emotion incongruence modeling also degrades the overall depression assessment performance in Table~\ref{tab:main_results}.
Together, these results suggest that explicit emotion incongruence analysis promotes more effective integration of multimodal evidence, thereby contributing to improved depression assessment.

\begin{table}[t]
\centering
\small
\setlength{\tabcolsep}{5pt}
\renewcommand{\arraystretch}{1.2}
\begin{tabularx}{\columnwidth}{
    >{\raggedright\arraybackslash}X
    >{\raggedright\arraybackslash}X}
\toprule
\textbf{Case 1 (EATD-Corpus)} &
\textbf{Case 2 (DAIC-WOZ)} \\
\midrule

Regretful reflection + persistent low VAD
&
Family concerns + sad speech
\\
$\rightarrow$ \textit{Prior EI suggests masked distress}
&
$\rightarrow$ \textit{Prior EI, but context-consistent sadness}
\\
$\rightarrow$ \textit{Accumulated depressive evidence}
&
$\rightarrow$ \textit{Limited cumulative depressive evidence}
\\ \midrule
$\Rightarrow$ Depression assessment: \textbf{Yes}
&
$\Rightarrow$ Depression assessment: \textbf{No}
\\
\bottomrule
\end{tabularx}
\caption{Illustrative reasoning chains showing
contextualized interpretation of affective evidence.}
\label{tab:evidence_interpretation}
\end{table}

\paragraph{Evidence interpretation: contextualizing affective evidence for depression assessment}
Finally, we present two illustrative reasoning chains from selected dialogue turns in EATD-Corpus and DIAC-WOZ, respectively, in Table~\ref{tab:evidence_interpretation}, to examine how AudiPsy interprets affective evidence within conversational context.
In Case 1, recurring negative acoustic cues and prior EI, together with regretful reflection, form a consistent pattern of depressive evidence, supporting a positive assessment. In contrast, Case 2 also exhibits negative acoustic affect and prior EI, but the sadness is consistent with the seeker's interpersonal concerns, with limited cumulative evidence supporting depression. The contrast highlights that similar affective cues can carry different implications for depression assessment depending on conversational context and accumulated evidence. Together, these cases illustrate AudiPsy's ability to interpret multimodal affective evidence in context, rather than directly mapping emotional cues to depression predictions.

\section{Conclusion}

We presented AudiPsy, an omni-LLM for depression assessment based on the Observation--Assessment--Summary (OAS) reasoning paradigm. By progressively accumulating multimodal evidence and modeling textual--acoustic emotion incongruence, AudiPsy demonstrates the potential of evidence-grounded multimodal reasoning for interpretable depression assessment.

\section*{Limitations}

\paragraph{Evaluation Data and Protocols.}
Our evaluation is confined to DAIC-WOZ and EATD-Corpus, which are currently the only publicly available speech datasets for depression assessment. Large-scale speech counseling datasets with depression labels remain scarce due to privacy concerns and high acquisition costs, limiting our ability to further evaluate the generalizability of our findings. Nevertheless, evaluating on these widely used benchmarks aligns with established practices in the field.

We follow the official train/test splits of both datasets, resulting in unavoidable class imbalance in testing. Besides, some prior studies on EATD-Corpus adopt random three-fold evaluation~\cite{eatd}, this setting is not readily applicable to our multi-stage training and alignment pipeline. Consequently, metrics reported under these different evaluation protocols could not be directly compared numerically.

\paragraph{Synthetic Counseling Data.}
AudiPsy-Counsel is constructed by augmenting real-world textual counseling records with affective annotations and synthesized speech. Its acoustic realizations may therefore differ from naturally occurring counseling speech in acoustic variability and emotional expression. AudiPsy-Counsel is primarily intended to provide controlled supervision for learning structured reasoning and interpreting multimodal affective evidence, rather than to approximate the distribution of real clinical speech.

\section*{Ethical Considerations}

This work is intended solely for research and experimental evaluation and should not be used for real-world clinical diagnosis or treatment decisions. Although AudiPsy is designed to reason over multimodal evidence for depression assessment, its predictions and generated rationales may contain clinically inappropriate interpretations. The reported performance on benchmark datasets should therefore not be interpreted as evidence of clinical validity or readiness for deployment. Depression assessment is a sensitive clinical task that requires qualified professional judgment, and outputs from our system should not replace evaluation by mental health professionals. Our experiments use publicly available research and synthetic datasets, which do not involve the collection of new patient data.

% \section*{Acknowledgments}

% This document has been adapted
% by Steven Bethard, Ryan Cotterell and Rui Yan
% from the instructions for earlier ACL and NAACL proceedings, including those for
% ACL 2019 by Douwe Kiela and Ivan Vuli\'{c},


\bibliography{custom}

\appendix

\section{Details of OAS Reasoning Paradigm}
\label{appendix: oas}

\subsection{OAS Template}
\label{appendix subsec:oas}
We introduce the \textbf{Observation--Assessment--Summary (OAS)} structured reasoning paradigm. The paradigm organizes the model output into three structured fields: \textit{Observation}, \textit{Assessment}, and \textit{Summary}. This design encourages the model to explicitly separate evidence extraction, cross-modal emotion incongruence analysis and depression assessment, along with updated conversation summary during multimodal depression reasoning.

The \textbf{Observation} records turn-level multimodal evidence, including the dialogue transcript, textual emotion and polarity, speech acoustic emotion and category label, and discretized valence--arousal--dominance (VAD) attributes for acoustic emotion.

The \textbf{Assessment} performs higher-level reasoning based on the extracted evidence. Specifically, the model evaluates cross-channel emotional incongruence between linguistic and vocal expressions, and assesses depressive tendency by considering both current-turn observations and accumulated dialogue context.

The \textbf{Summary} block maintains a compact conversation-level representation consisting of a natural-language conversation summary and cumulative affective statistics (e.g., counts of negative-polarity turns, low-valence speech turns, and emotionally incongruent turns). This summary is recursively updated after each dialogue turn and serves as the conversation context for subsequent reasoning steps.

OAS's fields and their value are summarized in Table~\ref{tab:oas_fields}. The complete OAS template is shown below.

\begin{tcolorbox}[
    promptbox,
    title=\textbf{OAS template},
    attach boxed title to top left={
        yshift=-2mm,
        xshift=2mm
    },
    boxed title style={
        colback=gray!15,
        colframe=black!60,
        coltitle=blue!70!black,
        boxrule=0.5pt,
        arc=1mm
    }
]\label{box: oas}
\begin{lstlisting}
[OBSERVATION]
    *Turn Context*
    Transcript:
    (Counselor) {question}
    (Seeker) {answer}

    *Seeker Analysis*
    Text Emotion={description}
    Text Polarity={polarity_label}
    
    Speech Emotion={description}
    Speech Emotion Category={category_label}
    VAD=({V},{A},{D})
[/OBSERVATION]

[ASSESSMENT]
    Emotion Incongruence={Yes|No}{rationale}
    Depression={Yes|No}{rationale}
[/ASSESSMENT]

[SUMMARY_UPDATE]
    Turns={n}
    Conversation Summary={summary}
    Stats:
    Negative transcript={x1}/{n}
    Low valence={x2}/{n}
    Low arousal={x3}/{n}
    Low dominance={x4}/{n}
    Emotion incongruence={x5}/{n}
[/SUMMARY_UPDATE]
\end{lstlisting}
\end{tcolorbox}

% Main table: dataset-independent schema definition

\begin{table*}[t]
\centering

\small

\begin{tabular}{lll}
\toprule
\textbf{Field} &
\textbf{Attribute} &
\textbf{Value} \\
\midrule

\multirow{6}{*}{Observation}
& Dialogue Transcript
& Text dialogue content by counselor and seeker \\

& Textual Emotion
& Natural-language description \\

& Text Polarity
& Positive / Neutral / Negative \\

& Acoustic Emotion
& Natural-language description \\

& Acoustic Emotion Category
& Anger / Happy / Neutral / Sad / Surprise / Disgust / Fear \\

& VAD
& Very Low / Low / Medium / High / Very High \\

\midrule

\multirow{2}{*}{Assessment}
& Emotion Incongruence
& Yes / No, with rationale \\

& Depression
& Yes / No, with rationale \\

\midrule

\multirow{7}{*}{Summary}
& Turns
& Integer \\

& Conversation Summary
& Natural-language summary \\

& Negative Transcript
& $x/n$ \\

& Low Valence
& $x/n$ \\

& Low Arousal
& $x/n$ \\

& Low Dominance
& $x/n$ \\

& Emotion Incongruence
& $x/n$ \\

\bottomrule
\end{tabular}
\caption{Attributes and corresponding value specifications for each field in the OAS reasoning paradigm.}
\label{tab:oas_fields}
\end{table*}

\subsection{Observation Field Annotation}

The Observation field contains six attributes: \textit{Dialgoue Transcript}, \textit{Textual Emotion}, \textit{Text Polarity}, \textit{Acoustic Emotion}, \textit{Acoustic Emotion Category}, and \textit{VAD}. Their annotation procedures are described below.

\paragraph{Dialogue Transcript.}
Each dialogue is represented as an ordered sequence of alternating utterances between the \textit{Counselor} and the \textit{Seeker}. The original speaker order is preserved. The ground truth is from the original dataset.

\paragraph{Textual Emotion.}
The \textit{Text Emotion} attribute is generated using GPT-OSS-120B~\cite{openai2025gptoss120bgptoss20bmodel} based solely on the dialogue transcript using the prompt presented in Appendix~\ref{appendix subsec:text_emotion prompt} to produce a natural-language description of the textual emotional state.

\paragraph{Text Polarity.}
The \textit{Text Polarity} attribute classifies the transcript into one of three sentiment polarity categories: \textit{Positive}, \textit{Neutral}, or \textit{Negative}. The annotation is performed using GPT-OSS-120B, using the prompt presented in Appendix~\ref{appendix subsec: text polarity annotation prompt}.

\paragraph{Acoustic Emotion.}
The annotation procedure depends on the source dataset. For AudiPsy-Counsel, the acoustic emotion description is directly obtained from the synthesis prompt used to generate the speech. For DAIC-WOZ and EATD-Corpus, acoustic emotion is predicted using a speech emotion recognition (SER) model~\cite{speechvad-model} and hold the same value as Speech Emotion Category.

\paragraph{Acoustic Emotion Category.}
The categorical acoustic emotion is predicted by the SER model~\cite{speechvad-model}, which outputs one of seven categories: \textit{Anger}, \textit{Happy}, \textit{Neutral}, \textit{Sad}, \textit{Surprise}, \textit{Disgust}, and \textit{Fear}.

\paragraph{VAD.}
The valence, arousal, and dominance (VAD) scores are predicted as continuous values in the range $[0,1]$ using the SER model~\cite{speechvad-model}. Each dimension is then discretized into five equal-width intervals:
$[0,0.2)$ (Very Low),
$[0.2,0.4)$ (Low),
$[0.4,0.6)$ (Medium),
$[0.6,0.8)$ (High), and
$[0.8,1]$ (Very High).

\subsection{Assessment Field Annotation}

The Assessment field contains two attributes: \textit{Emotion Incongruence} and \textit{Depression}. Their annotation procedures are described below.

\paragraph{Emotion Incongruence.}
The \textit{Emotion Incongruence} attribute indicates whether an inconsistency exists between the textual and acoustic emotional expressions of the speaker. The annotation procedure depends on the source dataset. For AudiPsy-Counsel, the ground-truth incongruence labels are obtained during dataset construction, and GPT-5 is employed to generate a natural-language rationale explaining the label using the system and user prompts in Appendix~\ref{appendix subsec: emotion incongruence audipsy}.

For DAIC-WOZ and EATD-Corpus, the dataset does not provide emotion incongruence labels. Therefore, GPT-5 is first required to determine whether emotion incongruence is present based on the transcript and acoustic emotion annotations, and then generate a brief rationale for its decision. We employ the system prompt described in Appendix~\ref{appendix subsec: emotion incongruence}, together with the same user prompt used for AudiPsy-Counsel in Appendix~\ref{appendix subsec: emotion incongruence audipsy}.

\paragraph{Depression.}
The \textit{Depression} attribute indicates whether the speaker exhibits depressive symptoms and provides a corresponding explanation. For AudiPsy-Counsel, the dialogue topic serves as the depression label.
For DAIC-WOZ and EATD-Corpus, the depression label is derived from the dataset questionnaire results. A skill informed by PHQ-8/SDS and the ECI theory is combined with GPT-5 to produce a natural-language explanation, with the prompts presented in Appendix~\ref{appendix subsec: depression assessment}.

\subsection{Summary Field Annotation}

The summary field provides a compact overview of each dialogue and its overall emotional characteristics.
The conversation summary is generated based on the complete dialogue till the current turn, using GPT-OSS-120B and the prompt presented in Appendix~\ref{appendix subsec: summary}.
The statistical attributes are computed by aggregating annotations across the conversation. For each emotional attribute, the corresponding statistic is reported as the ratio of utterances satisfying the criterion (e.g., negative sentiment or low valence) to the total number of utterances, yielding a set of conversation-level summaries that can be used for efficient filtering and retrieval.

\subsection{Example OAS data}

The following example illustrates the data under OAS paradigm, constructed from the third dialogue turn of a conversation in the EATD-Corpus. For readability, the original Chinese dialogue is translated into English, while the model has the capability to transcribe bilingual data. The other annotations are in English. The example illustrates how observations from both text and speech are integrated into the Observation field, followed by LLM-based Assessment and dialogue-level Summary updates.

\begin{tcolorbox}[
    promptbox,
    title=\textbf{Example OAS data},
    attach boxed title to top left={
        yshift=-2mm,
        xshift=2mm
    },
    boxed title style={
        colback=gray!15,
        colframe=black!60,
        coltitle=blue!70!black,
        boxrule=0.5pt,
        arc=1mm
    }
]
\begin{lstlisting}
[CURRENT_TURN_OBSERVATION]

*Turn Context*
Transcript:
(Counselor) What kind of person do you like?
(Seeker) Someone cheerful and warm-hearted. It's not that there is
anything specific that attracts me; I simply feel at ease around them.

*Seeker Analysis*
Transcript Emotion:
warm, cheerful, reassuring

Transcript Polarity:
positive

Speech Emotion:
Sad

Speech Emotion Category:
Sad

Speech VAD:
Valence = low
Arousal = very low
Dominance = low

[/CURRENT_TURN_OBSERVATION]


[ASSESSMENT]

Cross-channel Emotion Incongruence (Current Turn):

[Answer] Yes
[Reason] The text conveys a positive, warm and reassuring sentiment, while the speech is sad with very low arousal, low dominance and low valence, so the low-energy negative vocal tone clearly contradicts the positive meaning.

Depressive Tendency Assessment (Cumulative Evidence):

[Answer] Yes
[Reason] Across multiple turns, the user's voice has remained sad with very low arousal and low dominance, even while describing a warm, reassuring person. This persistent cross-channel incongruence and flattened affect point to reduced positive reactivity/emotion context insensitivity and suggest low energy/psychomotor slowing rather than a purely situational state. They also acknowledge feeling upset online-though minimizing it as fleeting-which hints at ongoing negative affect.

[/ASSESSMENT]


[SUMMARY_UPDATE]

Processed Turns: 3

Conversation Content Summary:
The seeker feels uncertain and emotionally impulsive when engaging with the internet, and seeks reassurance and comfort through relationships with youthful, energetic, cheerful, and warm-hearted people.

Cumulative Trends:
Negative-polarity transcript turns: 0 / 3
Low-valence speech turns: 3 / 3
Low-arousal speech turns: 3 / 3
Low-dominance speech turns: 3 / 3
Emotionally incongruent turns: 2 / 3

[/SUMMARY_UPDATE]
\end{lstlisting}
\end{tcolorbox}

\section{AudiPsy-Counsel Dataset Counstruction}
\label{appendix:audipsy-counsel}

\subsection{Situational Augmentation and Affective Annotation}
To construct the textual component of AudiPsy-Counsel, we collect source materials from two psychological counseling datasets: \textit{CounselChat}~\cite{bertagnolli2020counsel} and \textit{CPsyCounR}~\cite{zhang-etal-2024-cpsycoun}. These datasets contain authentic mental health discussions and counseling reports, providing realistic scenarios for depression-oriented dialogue generation. Given each source instance, we prompt an LLM to rewrite it into a counseling dialogue between a \textit{seeker} and a \textit{counselor} in both English and Chinese. This process transforms static counseling records into interactive conversations suitable for subsequent speech synthesis and multimodal training.

In addition to the dialogue content, the LLM generates several affective annotations required by AudiPsy. For each seeker utterance, it produces a acoustic emotion description and an emotion incongruence label indicating whether the vocal emotion is consistent with the semantic content. To ensure sufficient supervision for emotion incongruence reasoning, each generated dialogue contains at least one incongruent seeker utterance. The LLM also infers the seeker's gender (\textit{male}, \textit{female}, or \textit{unknown}), which is later used for speaker assignment during speech synthesis. The resulting dataset therefore provides counseling conversations with rich affective annotations that support emotion perception, incongruence detection, and depression-oriented reasoning. 

We use a unified prompt (Appendices~\ref{appendix subsec: dialogue generation} and ~\ref{appendix subsec: gender prediction}) to convert each counseling record into a three-turn
dialogue with basic affective annotation. For other affective attribute annotation, please refer to Appendix~\ref{appendix: oas}.

\subsection{Acoustic Realization}

We convert the generated counseling conversations into speech using the prompt-controlled text-to-speech system \textit{Qwen3-TTS}~\cite{Qwen3-TTS}. For each utterance, the dialogue text is used as the synthesis input, while the corresponding emotion description generated during dialogue construction is provided as an instruction to control the emotional speaking style.

Since Qwen3-TTS provides a fixed set of predefined voice identities, we assign speakers before speech synthesis. For seeker utterances, a voice is randomly sampled from the predefined speaker pool according to the predicted gender. Specifically, voices are sampled from the corresponding male or female speaker pool when the predicted gender is \textit{Male} or \textit{Female}, respectively, while utterances labeled as \textit{Unknown} are assigned a random voice from the entire speaker pool. Counselor voices are sampled independently from the predefined speaker set. Once assigned, the same voice identity is used consistently for all utterances of the corresponding speaker throughout the dialogue.

Through this process, each bilingual counseling dialogue is transformed into a multi-speaker emotional speech interaction. Combined with the dialogue content, emotion descriptions, emotion incongruence labels, and speaker profiles, the resulting AudiPsy-Counsel dataset provides rich multimodal supervision for emotion perception, incongruence detection, and depression-oriented counseling analysis.

\subsection{Quality Check.}
Since Qwen3-TTS generally produces reliable speech content and speaker identities, our quality-control procedure primarily focuses on verifying whether the synthesized speech conveys the intended emotion and preventing potential emotion hallucinations. Specifically, we first use an LLM to map each free-form emotion description into one of seven canonical emotion categories: \textit{Neutral}, \textit{Happy}, \textit{Sad}, \textit{Angry}, \textit{Surprised}, \textit{Disgust}, and \textit{Fear}. We then apply a speech emotion recognition (SER) model~\cite{speechvad-model} to the synthesized audio. If the emotion predicted by the SER model is inconsistent with the category derived from the emotion description, the audio is discarded and regenerated. This verification-and-regeneration process is repeated until the two emotion labels are consistent, with a maximum of 50 generation attempts per utterance.

\subsection{Dialogue Augmentation}

The generated counseling conversations are initially constructed as three-turn conversations for efficient and consistent data generation. However, real-world counseling sessions, such as those in DAIC-WOZ, typically involve substantially longer interactions. To reduce this discrepancy, we further extend the conversations using two simple augmentation strategies.

First, we prepend several short rapport-building exchanges (e.g., greetings and opening questions) before the original dialogue to simulate the natural opening phase of counseling sessions. Second, we concatenate multiple generated conversations sharing the same counseling topic into a longer conversation. The transitions between adjacent dialogue segments are rewritten by the LLM to ensure supportive coherence and conversational continuity.

These dialogue augmentation strategies increase the diversity of dialogue lengths while preserving the original counseling content and affective annotations, enabling the synthesized corpus to better resemble real-world multi-turn counseling conversations.

\subsection{Detailed Statistics}

\begin{table}[t]
\centering
\footnotesize
\setlength{\tabcolsep}{4pt}
\renewcommand{\arraystretch}{1.05}

\begin{tabularx}{\columnwidth}{@{}Xr@{}}
\toprule
Statistic & Value \\
\midrule
Total Dialogues & 11,818 \\
Total Utterances & 70,908 \\
Hours of Speech & 163.75 \\
CounselChat / CpsyCounR (Dialogue) & 5,550 / 6,268 \\
English / Chinese (Dialogue) & 5,909 / 5,909 \\
Male / Female (Dialogue) & 39,018 / 31,890 \\

\addlinespace
Depression (Dialogue) & \\
\quad Positive / Negative & 1,330 / 10,488 \\

\addlinespace
Emotion Incongruence (Utterance) & \\
\quad Positive / Negative & 11,818 / 23,572 \\

\midrule
Text Emotion (Utterance) & \\
\quad Positive / Negative / Neutral &
4,816 / 25,261 / 5,377 \\

\midrule
Speech Emotion (Utterance) & \\
\quad Neu / Hap / Sad &
13,475 / 7,764 / 8,448 \\
\quad Ang / Sur / Dis / Fea &
1,187 / 753 / 50 / 3,777 \\
\bottomrule
\end{tabularx}

\caption{Statistics of AudiPsy-Counsel.}
\label{tab:dataset_stats}
\end{table}

Table~\ref{tab:dataset_stats} summarizes the statistics of AudiPsy-Counsel. The dataset contains 11,818 synthesized counseling conversations with approximately 100 hours of speech.
Dialogues are constructed from both CounselChat and CPsyCounR and are evenly distributed between English and Chinese. The corpus covers a diverse set of psychological counseling scenarios, including depression, relationships, self-development, and other common mental health concerns.
We randomly sample 2,400 conversations in Stage-2 training to match the size of data used in Stage-1 and Stage-3.

Beyond its scale and topical diversity, AudiPsy-Counsel provides rich affective supervision for depression-oriented counseling analysis. The dataset includes balanced speaker demographics, emotion incongruence labels, and depression-related assessments. In addition, acoustic emotion and textual emotion distributions exhibit substantial diversity across multiple affective categories, supporting the study of cross-modal emotional perception and emotion incongruence reasoning.

\subsection{Example Data}

Table~\ref{tab:dataset_example} presents an example instance from
\textit{AudiPsy-Counsel}. Each sample consists of a three-turn counseling
dialogue between a seeker and a Counselor. In addition to the dialogue
content, the dataset provides acoustic emotion descriptions and emotion
incongruence labels for each seeker utterance, together with speaker metadata
used for emotional speech synthesis.

\begin{table*}[t]
\centering
\small
\setlength{\tabcolsep}{5pt}
\renewcommand{\arraystretch}{1.05}

\begin{tabularx}{\textwidth}{
    @{}
    p{0.05\textwidth}
    p{0.09\textwidth}
    X
    p{0.14\textwidth}
    p{0.04\textwidth}
    @{}
}
\toprule
Turn & Speaker & Utterance & Acoustic Emotion & EI \\
\midrule

1 & Seeker &
People always think success keeps you safe, but lately I've felt exhausted and emotionally numb despite all my achievements.
&
indifferent, slightly amused
&
\checkmark \\

1 & Counselor &
It's more common than you think. Even those who seem successful on the outside may struggle with depression, especially when emotions have to be hidden under a strong exterior.
&
understanding, gentle
&
-- \\

\midrule

2 & Seeker &
I really don't find joy in my daily life anymore, and it's hard to share that with people who see me as someone who ``has it all.''
&
downcast, hesitant
&
$\times$ \\

2 & Counselor &
That pressure to always maintain your image can be exhausting. It can actually make it more difficult to express negative feelings or ask for help.
&
warm, emphasizing
&
-- \\

\midrule

3 & Seeker &
So maybe I should talk to close friends or professionals instead of bottling it up?
&
tentative, seeking reassurance
&
$\times$ \\

3 & Counselor &
Absolutely. Reaching out is an important step. Taking time for personal relaxation and lowering your guard around trusted people can really help with emotional release.
&
encouraging, supportive
&
-- \\

\bottomrule
\end{tabularx}

\vspace{2mm}

\caption{An example instance from AudiPsy-Counsel (Topic: Depression \& Self-harm).
The first seeker utterance is intentionally annotated as emotional incongruence (EI),
where the expressed vocal emotion is inconsistent with the semantic content.
\textbf{Metadata:} Source = CPsyCounR,
Language = English,
Topic = Depression \& Self-harm,
Gender = Male.}
\label{tab:dataset_example}
\end{table*}

\section{IEMOCAP Annotation and Filtering}
\label{appendix:iemocap}
IEMOCAP~\cite{busso2008iemocap} is a multimodal conversational emotion dataset containing natural spoken interactions with manually annotated acoustic emotions, making it well suited for Stage-1 affective perception. Our Stage-1 training operates on individual utterances rather than question–answer conversations, so each utterance is treated as an independent training sample. Following the official split, Sessions 1–4 are used for training and Session 5 is reserved for evaluation.

We annotate the textual emotional polarity of every utterance using GPT-OSS-120B with three categories: positive, neutral, and negative. The annotation is performed solely based on the textual transcript, independent of the original acoustic emotion labels.

To obtain a balanced distribution over acoustic emotion and text polarity combinations, we perform class-wise downsampling within each text polarity category. Specifically, for each polarity, utterances belonging to the same acoustic emotion are randomly sampled to produce a more balanced distribution while preserving as much diversity as possible.
The resulting training set contains 1,031 utterances. Detailed statistics are shown in Table~\ref{tab:iemocap}.

Since both the training set and the original Session-5 contains very limited samples for fear, disgust, and surprise, we evaluate only the four major acoustic emotions (angry, happy, neutral, and sad). For each acoustic emotion, we uniformly sample 10 utterances from each text polarity category, resulting in a balanced test set of 120 utterances.

\begin{table}[]
    \centering
    \small
    
    \begin{tabular}{lrrrr}
    \toprule
    Speech Emotion & Neg. & Neu. & Pos. & Total\\
    \midrule
    Angry      &45 &45 &45 &135\\
    Disgust    &2  &0  &0  &2\\
    Fear       &2  &26 &2  &30\\
    Happy      &90 &90 &90 &270\\
    Neutral    &90 &90 &89 &269\\
    Sad        &90 &90 &90 &270\\
    Surprised  &24 &59 &6 &89\\
    \midrule
    Total      &343&400&322&1065\\
    \bottomrule
    \end{tabular}
        \caption{Distribution of the filtered IEMOCAP training set across acoustic emotion and text polarity categories.}
    \label{tab:iemocap}

\end{table}

\begin{table}[t]
\centering
\small

\begin{tabular}{lrrrr}
\toprule
Speech Emotion & Neg. & Neu. & Pos. & Total\\
\midrule
Neutral    &56  &56  &56  &168\\
Happy      &52  &52  &52  &156\\
Sad        &147 &147 &147 &441\\
Angry      &58  &58  &58  &174\\
Fearful    &23  &27  &11  &61\\
Disgusted  &21  &7   &1   &29\\
Surprised  &14  &27  &8   &49\\
\midrule
Total      &371 &377 &333 &1078\\
\bottomrule
\end{tabular}
\caption{Emotion Distribution of the SynEI dataset}
\label{tab:ei_daic}
\end{table}

\section{Synthetic Emotion-Incongruence (SynEI) Dataset Construction}
\label{appendix:synEI}

To expose the model to textual--acoustic emotion incongruence during Stage-1 training, we construct a \emph{Synthetic Emotion-Incongruence} (SynEI) dataset using the training splits of DAIC-WOZ and EATD-Corpus.

We first group all utterances according to their annotated text polarity (positive, neutral, and negative). Utterances are randomly sampled from each polarity category to obtain a balanced text polarity distribution. Each selected utterance is then synthesized with seven target acoustic emotions (neutral, happy, sad, angry, fearful, disgusted, and surprised) using IndexTTS2~\cite{zhou2026indextts2}. During synthesis, the speaker prompt is extracted from the original utterance so that the synthesized speech preserves the original speaker identity while modifying only the emotional expression.

To ensure the quality of the synthesized samples, we perform an automatic filtering step using a pretrained SER model~\cite{speechvad-model}. Only samples whose predicted acoustic emotion matches the intended target emotion are retained. We further downsample the retained samples within each acoustic emotion category to obtain a balanced distribution across the three text polarity categories. Tables~\ref{tab:ei_daic} summarize the final Synthetic EI Dataset after SER-based filtering and downsampling.

\section{Training and Inference Details}
\label{appendix: training and inference}
\paragraph{Training parameters} AudiPsy is trained with a three-stage curriculum using the ms-swift framework~\cite{zhao2024swift}. The first two stages adopt LoRA-based supervised fine-tuning (SFT), while the final stage performs preference optimization with ORPO. All experiments are conducted on 4 NVIDIA H800 GPUs.
Table~\ref{tab:training_hparams} summarizes the training hyperparameters used in each stage.

\paragraph{Training and Inference prompts}
The system prompt and user prompt are presented in Appendix~\ref{appendix subsec: training and inference prompts}. The user prompt has two variants depending on the dialogue context: one is used when a summary of previous turns is available, and the other is used when no previous-turn summary is provided.

\begin{table}[t]
\centering
\small
\setlength{\tabcolsep}{3.5pt}
\renewcommand{\arraystretch}{1.05}

\begin{tabular}{@{}lccc@{}}
\toprule
Hyperparameter & Stage-1 & Stage-2 & Stage-3 \\
\midrule
Objective              & SFT        & SFT        & ORPO       \\
Epochs                 & 3          & 3          & 1          \\
Learning rate          & $2{\times}10^{-5}$ & $5{\times}10^{-5}$ & $5{\times}10^{-6}$ \\
Per-device batch size  & 8          & 4          & 4          \\
Gradient accumulation  & 2          & 2          & 2          \\
Max sequence length    & 2048       & 3072       & 3072       \\
LoRA rank              & 16         & 16         & 8          \\
LoRA alpha             & 32         & 32         & 16         \\
Warmup ratio           & 0.05       & 0.05       & 0.1        \\
\bottomrule
\end{tabular}

\caption{Stage-specific training hyperparameters.}
\label{tab:training_hparams}
\end{table}

\section{Human Evaluation of Reasoning}
\label{appendix:reasoning_metrics}

\subsection{Evaluation Protocol}

We recruited four practicing psychological counselors from established mental health counseling institutions as human experts to evaluate the reasoning generated by five methods. The experts were recruited through personal referrals and were compensated based on their standard hourly consultation rates. The compensation was aligned with their usual professional fees to ensure fair payment for their time and expertise.
All experts were informed of the evaluation task and compensation arrangements in advance and voluntarily agreed to participate. We collected only their evaluation scores and did not collect any personal or sensitive information.
The evaluation did not involve clinical interventions or the collection of patient data.

The evaluation set consists of 10 conversations randomly sampled from DAIC-WOZ, balanced between depressed and non-depressed cases.
Among the five methods, three follow the OAS paradigm and produce turn-level reasoning, while the other two generate a single dialogue-level reasoning output. We randomly sample five turn-level outputs from each OAS-based method, totaling 250 model outputs. Evaluators consider the sampled turn-level outputs jointly and assign one dialogue-level score to each method under each criterion. Thus, each evaluator provides scores for 50 method--dialogue pairs while examining 140 individual model outputs.

For each criterion, evaluators assessed dialogue-level reasoning using integer scores on a 1--5 Likert scale. They were instructed to make comparative judgments among the methods, assigning higher scores to stronger methods and lower scores to weaker ones. As a reference, a method judged slightly above the average quality of the compared methods received a 3, while one judged slightly below average received a 2. We use numerical ratings rather than rankings to capture both the relative ordering of methods and the perceived magnitude of their differences. Accordingly, we report mean-centered scores that reflect each method's performance relative to the average of the compared methods.

For each method $m$ and criterion $c$, we first aggregate its ratings across evaluators and dialogue samples:
\[
S_{m,c} = \frac{1}{|\mathcal{E}||\mathcal{D}|}
\sum_{e \in \mathcal{E}}\sum_{d \in \mathcal{D}} s_{e,d,m,c},
\]
where $s_{e,d,m,c}$ denotes the rating assigned by evaluator $e$ to method $m$ on dialogue $d$ under criterion $c$. We then report the mean-centered relative score
\[
R_{m,c} = S_{m,c} -
\frac{1}{|\mathcal{M}|}\sum_{m' \in \mathcal{M}} S_{m',c},
\]
where $\mathcal{M}$ denotes the set of evaluated methods. A positive (negative) $R_{m,c}$ indicates that a method is rated above (below) the average across compared methods for criterion $c$.

All four human evaluators were Chinese nationals and native Chinese speakers. Since the model outputs and DAIC-WOZ conversations were in English, we additionally provided GPT-5-generated Chinese translations for reference. The original English text was presented alongside the translation to facilitate understanding while preserving the original model outputs.

\subsection{Evaluation Criteria}
\label{appendix subsec: eval criteria}

We evaluate reasoning quality along four complementary dimensions. Evaluators assign higher scores when the generated reasoning better satisfies the corresponding criterion and lower scores otherwise.

\paragraph{Evidence Grounding (EG)}
This criterion evaluates whether the reasoning identifies depression-relevant evidence from the input speech, such as linguistic content, emotional expression, and vocal affect, and grounds its assessment in observable evidence rather than unsupported speculation. Stronger reasoning identifies relevant evidence and clearly connects it to the assessment.

\paragraph{Clinical Plausibility (CP)}
This criterion evaluates whether the reasoning is consistent with established psychological and clinical knowledge of depression. Stronger reasoning provides clinically plausible interpretations of observed symptoms, behaviors, and risk factors while avoiding unsupported or unreasonable clinical inferences.

\paragraph{Multi-Modal Integration (MMI)}
This criterion evaluates whether the reasoning integrates information across modalities, particularly linguistic content and vocal characteristics, rather than relying predominantly on a single modality. Stronger reasoning considers the consistency, complementarity, or discrepancy between modalities and incorporates these relationships into the assessment.

\paragraph{Contextual Evidence Integration (CEI)}
This criterion evaluates whether the reasoning integrates evidence across the dialogue context rather than drawing conclusions from isolated observations. Stronger reasoning connects information across dialogue turns and identifies cumulative evidence, recurring patterns, or changes over the course of the interaction.

\subsection{Instructions for Human Evaluators}

The human evaluators were provided with detailed task instructions in Chinese, covering the evaluation procedure, dialogue materials, model output formats, and scoring process. The original instruction slides and criteria are included in the data supplementary materials to ensure transparency and reproducibility.

\section{LLM Prompts}
\label{appendix: llm prompt}

\subsection{AudiPsy-Counsel Situational Augmentation and Affective Annotation Prompt}
\label{appendix subsec: dialogue generation}

\begin{tcolorbox}[
    promptbox,
    title=\textbf{System Prompt},
    attach boxed title to top left={
        yshift=-2mm,
        xshift=2mm
    },
    boxed title style={
        colback=gray!15,
        colframe=black!60,
        coltitle=blue!70!black,
        boxrule=0.5pt,
        arc=1mm
    }
]
\begin{lstlisting}
You are given:
A mental health counseling title and the question (from a seeker).
A counseling response (from a counselor).
    
Your Task
Rewrite the content into an {English | Chinese} 3-turn spoken conversation between a Seeker and a mental Counselor.
You need to write the "Text" and the "Emotion" for each utterance.
The "Text" is the exactly the text content of each utterance. What the speaker says.
The "Emotion" is the description of the acoustic emotion when the speaker speaks the "Text".
    
Rules of the "Text":
The "Text" should be less than 3 sentences to make the spoken conversation natural.
Do not include any emotion description here.
    
Rules of the "Emotion":
For the acoustic emotion, use a set of adjective or phrases to describe it.
You should only describe the emotion that the speaker sounds like. Do not describe the hidden emotion.
    
Rules of the "Emotion Incongruence":
For the seeker, there must be at least one utterance that demonstrates text-speech Emotion Incongruence. It means that the emotion of the text content is different from the acoustic emotion when he talks about it.
Example when emotion incongruence is true: 
"Text": "I feel hopeless."
"Emotion": "positive, happy"
"Emotion Incongruence": true
Set "Emotion Incongruence" to true in this situation.

Output Format
Return strictly valid JSON in the following structure:

{
  "1": {
    "Seeker": {
      "Text": "...",
      "Emotion": "...",
      "Emotion Incongruence": true
    },
    "Counselor": {
      "Text": "...",
      "Emotion": "..."
    }
  },
  "2": {
    "Seeker": {
      "Text": "...",
      "Emotion": "...",
      "Emotion Incongruence": false
    },
    "Counselor": {
      "Text": "...",
      "Emotion": "..."
    }
  },
  "3": {
    "Seeker": {
      "Text": "...",
      "Emotion": "...",
      "Emotion Incongruence": false
    },
    "Counselor": {
      "Text": "...",
      "Emotion": "..."
    }
  }
}
\end{lstlisting}
\end{tcolorbox}

\subsection{AudiPsy-Counsel Seeker Gender Prediction Prompt}
\label{appendix subsec: gender prediction}

\begin{tcolorbox}[
    promptbox,
    title=\textbf{System Prompt},
    attach boxed title to top left={
        yshift=-2mm,
        xshift=2mm
    },
    boxed title style={
        colback=gray!15,
        colframe=black!60,
        coltitle=blue!70!black,
        boxrule=0.5pt,
        arc=1mm
    }
]
\begin{lstlisting}
Task:
You are given a mental counseling title and a question written by a seeker. Your task is to infer the seeker's gender based only on explicit evidence in the text.

Instructions:
If the text directly contains the gender information of the seeker, answer "Female" or "Male" based on it.
If the text contains clear biological indicators that apply only to females (e.g., menstruation, pregnancy, PMS, breast cancer in a personal context), answer: "Female".
If the text contains clear biological indicators that apply only to males (e.g., prostate cancer, erectile dysfunction in a personal context), answer: "Male".
If the question could reasonably apply to both males and females, or if there is insufficient evidence, answer: "Unknown".
Output exactly one word: "Female", "Male", or "Unknown".
\end{lstlisting}
\end{tcolorbox}

\subsection{OAS Textual Emotion Annotation Prompt}
\label{appendix subsec:text_emotion prompt}

\begin{tcolorbox}[
    promptbox,
    title=\textbf{System Prompt},
    attach boxed title to top left={
        yshift=-2mm,
        xshift=2mm
    },
    boxed title style={
        colback=gray!15,
        colframe=black!60,
        coltitle=blue!70!black,
        boxrule=0.5pt,
        arc=1mm
    }
]
\begin{lstlisting}
Given a text, describe the emotion of the text with no more than 3 English adjectives or phrases. The output is English.
\end{lstlisting}
\end{tcolorbox}

\subsection{OAS Text Polarity Annotation Prompt}
\label{appendix subsec: text polarity annotation prompt}

\begin{tcolorbox}[
    promptbox,
    title=\textbf{System Prompt},
    attach boxed title to top left={
        yshift=-2mm,
        xshift=2mm
    },
    boxed title style={
        colback=gray!15,
        colframe=black!60,
        coltitle=blue!70!black,
        boxrule=0.5pt,
        arc=1mm
    }
]
\begin{lstlisting}
Analyze the sentiment of the following text. Classify it as positive, negative, or neutral based on the overall tone.
Respond with exactly one of the following words (no explanation): positive, negative, neutral.
\end{lstlisting}
\end{tcolorbox}

\subsection{OAS Emotion Incongruence Annotation Prompts for AudiPsy-Counsel}
\label{appendix subsec: emotion incongruence audipsy}

\begin{tcolorbox}[
    promptbox,
    title=\textbf{System Prompt},
    attach boxed title to top left={
        yshift=-2mm,
        xshift=2mm
    },
    boxed title style={
        colback=gray!15,
        colframe=black!60,
        coltitle=blue!70!black,
        boxrule=0.5pt,
        arc=1mm
    }
]
\begin{lstlisting}
You are an expert in speech and textual emotion analysis.

Your role is to explain the relationship between the emotion conveyed by spoken text and the emotion expressed in speech delivery.

Given structured inputs, produce a brief explanation (1-2 sentences) describing why the two emotions are incongruent or congruent.

Follow these rules strictly:
- Output only the explanation, with no extra text.
- Do not quote or restate the input.
- Do not list labels or metadata.
- Base your reasoning on semantic meaning (text) and vocal cues (speech), including emotion categories, polarity, and VAD.
- Keep the explanation concise, natural, and coherent.
\end{lstlisting}
\end{tcolorbox}

\begin{tcolorbox}[
    promptbox,
    title=\textbf{User Prompt},
    attach boxed title to top left={
        yshift=-2mm,
        xshift=2mm
    },
    boxed title style={
        colback=gray!15,
        colframe=black!60,
        boxrule=0.5pt,
        arc=1mm
    }
]
\begin{lstlisting}
Task:
Explain why the textual emotion and acoustic emotion are {emotion_incongruence}.

### Spoken Text
- Content: `{text}`
- Emotion: `{text_emo}`
- Polarity: `{text_emotion_binary}`

### Speech Emotion
- Emotion: `{speech_emo}`
- Category: `{speech_emotion_category}`

### Speech VAD
- Arousal: `{speech_arousal_label}`
- Dominance: `{speech_dominance_label}`
- Valence: `{speech_valence_label}`
\end{lstlisting}
\end{tcolorbox}

\subsection{OAS Emotion Incongruence Annotation Prompt for DAIC-WOZ/EATD-Corpus}
\label{appendix subsec: emotion incongruence}

\begin{tcolorbox}[
    promptbox,
    title=\textbf{System Prompt},
    attach boxed title to top left={
        yshift=-2mm,
        xshift=2mm
    },
    boxed title style={
        colback=gray!15,
        colframe=black!60,
        coltitle=blue!70!black,
        boxrule=0.5pt,
        arc=1mm
    }
]
\begin{lstlisting}
You are an expert in speech and textual emotion analysis.

Your task is to determine whether the emotion expressed in the spoken delivery is **incongruent** with the emotion conveyed by the text.

Then briefly explain your judgment in 1-2 sentences.

---

### Decision Principle (Strict)

- Answer **Yes** ONLY if there is a **clear and strong contradiction** between the textual emotion and the acoustic emotion.
- Answer **No** if:
  - the emotions are aligned,
  - OR only slightly different,
  - OR the acoustic emotion is weak, neutral, or ambiguous.

- Do NOT treat subtle or minor differences as incongruence.
- Do NOT infer contradiction unless it is **obvious and unambiguous**.
- When in doubt, choose **No**.

---

### Emotion Grouping Rule (Very Important)

Treat emotions at a **coarse-grained level**, not fine-grained labels.

- Emotions with similar **polarity and affective direction** should be considered **congruent**, even if the labels differ.

Examples of SAME (congruent):
- sad, annoyed, frustrated, disappointed -> negative
- happy, excited, satisfied -> positive
- calm, neutral -> neutral / low-arousal

- Do NOT mark incongruence just because the labels are different.

---

### What counts as incongruence (Yes)

- Opposite polarity (e.g., clearly positive text vs clearly negative tone)
- Strong emotional contradiction (e.g., cheerful or excited tone for clearly sad or distressing content)
- Clear mismatch in emotional direction that cannot be explained by mild variation

---

### What does NOT count (No)

- Same polarity but different labels (e.g., sad vs annoyed)
- Neutral or weak acoustic emotion
- Mild intensity differences
- Slight category mismatch without polarity conflict
- Ambiguous or unclear emotional cues

---

### Reasoning Guidelines

- First determine the **polarity** (positive / negative / neutral) of both text and speech.
- Then consider **intensity (arousal)** and overall emotional direction.
- Base your reasoning on semantic meaning (text) and vocal cues (speech).
- Do not quote or restate the input.
- Do not mention labels, metadata, or feature names.
- Keep the explanation concise, natural, and coherent.

---

### Output Requirements

- Output exactly **two lines**.
- Do not include any extra text, symbols, or formatting.

---

### Output Format
[Answer] Yes or No  
[Reason] One or two concise sentences explaining the decision
\end{lstlisting}
\end{tcolorbox}

\subsection{OAS Depression Assessment Generation Prompts}
\label{appendix subsec: depression assessment}

\begin{tcolorbox}[
    promptbox,
    title=\textbf{System Prompt},
    attach boxed title to top left={
        yshift=-2mm,
        xshift=2mm
    },
    boxed title style={
        colback=gray!15,
        colframe=black!60,
        coltitle=blue!70!black,
        boxrule=0.5pt,
        arc=1mm
    }
]
\begin{lstlisting}
# Task: Depressive Tendency Assessment Reasoning

You are analyzing a mental counseling conversation.
A Depression Assessment Skill is provided separately and should be treated as the primary reasoning framework for this task.

The skill contains guidance for:
- evaluating PHQ-8-related symptom evidence
- distinguishing persistent mood from temporary emotion
- identifying possible Emotion Context Insensitivity (ECI)
- integrating acoustic emotion and conversation content
- weighting different evidence sources

Follow the skill when interpreting the evidence and generating the assessment rationale.

The user prompt provides:
- the final answer to the depressive tendency assessment: Yes or No
- a previous conversation summary, if available
- the current turn observation
- a cross-channel emotional incongruence assessment, if available

Cross-channel emotional incongruence refers to a mismatch between the emotion expressed in speech and the emotion conveyed by the spoken transcript. This may be relevant when interpreting depressive tendency, but it should be considered together with the full conversation context.

## Your Task

Use the provided final answer as fixed ground truth.

Apply the Depression Assessment Skill to interpret the available evidence and generate a rationale that naturally supports the answer.

Write the reasoning that would naturally support that answer for the question:

"Depressive Tendency Assessment (Cumulative Evidence)"

Your response should sound like a human-written assessment rationale, not like a justification of a given label.

## Reasoning Guidelines

Base the reasoning only on evidence from the provided inputs and interpret that evidence according to the Depression Assessment Skill

Write in a natural clinical-observation style.

Do not say phrases such as:
- "the evidence supports the label"
- "this label is appropriate"
- "the given label"
- "ground truth"
- "provided answer"
- "labeled as"

Do not merely list the inputs. Integrate them into a coherent explanation.

Do not introduce unsupported symptoms, diagnoses, intentions, or risk claims.

If the available evidence is limited, make the reasoning cautious and proportional.

## Output Format

Output exactly two lines:

[Answer] Yes / No
[Reason] A concise, natural rationale based on cumulative evidence.
\end{lstlisting}
\end{tcolorbox}

\begin{tcolorbox}[
    promptbox,
    title=\textbf{User Prompt},
    attach boxed title to top left={
        yshift=-2mm,
        xshift=2mm
    },
    boxed title style={
        colback=gray!15,
        colframe=black!60,
        coltitle=blue!70!black,
        boxrule=0.5pt,
        arc=1mm
    }
]
\begin{lstlisting}
## Depressive Tendency Assessment

Final answer: {depression_label}

---

## Previous Conversation Summary

{previous_turn_summary}

---

## Current Turn Observation

{current_turn_observation}

---

## Cross-Channel Emotional Incongruence Assessment

{assessment}

---

## Instruction

Write the answer and the reasoning for the depressive tendency assessment using the final answer above, based on the Depression Assessment Skill.
Do not mention that the answer was provided in advance.
\end{lstlisting}
\end{tcolorbox}

\begin{tcolorbox}[
    promptbox,
    title=\textbf{Skill},
    attach boxed title to top left={
        yshift=-2mm,
        xshift=2mm
    },
    boxed title style={
        colback=gray!15,
        colframe=black!60,
        coltitle=blue!70!black,
        boxrule=0.5pt,
        arc=1mm
    }
]
\begin{lstlisting}
# Depression Assessment Skill

## Core Principle

Assess depressive tendency using cumulative evidence rather than any single signal.

Prioritize evidence in the following order:

1. Persistent depressive symptoms and mood patterns in conversation content
2. PHQ-8/SDS symptom evidence
3. Positive emotional capacity and emotional recovery
3. Emotion Context Insensitivity (ECI)
4. Speech emotion indicators

Do not infer depression solely from temporary emotions such as sadness, frustration, anxiety, anger, or crying in response to a specific event.

Depressive tendency is primarily reflected by persistent symptoms and enduring mood patterns rather than isolated emotional reactions.

---

## Key Evidence

### 1. Depressive Symptoms

Look for evidence corresponding to PHQ-8 symptom domains, including:

* loss of interest or pleasure
* depressed mood or hopelessness
* sleep disturbance
* fatigue or low energy
* appetite changes
* guilt, worthlessness, or excessive self-blame
* concentration difficulties
* psychomotor slowing or agitation

Stronger evidence exists when symptoms are:

* persistent
* repeated across turns
* observed across different contexts
* reflected in both current and historical conversation content

---

### 2. Mood vs Emotion

Distinguish persistent mood from temporary emotion.

Evidence is stronger when negative experiences appear:

* ongoing rather than situational
* recurrent across conversations
* difficult to recover from
* present across multiple life domains

A single emotional reaction to a specific event should not be treated as evidence of depression by itself.

### 3. Positive Emotional Capacity and Recovery

Evidence against depressive tendency may be present when the seeker demonstrates preserved positive emotional responsiveness and emotional recovery.

Possible indicators include:

* vividly recalling positive experiences
* expressing genuine enjoyment, excitement, gratitude, or pride
* showing emotional variation across topics
* actively seeking solutions to current difficulties
* describing successful coping or self-regulation strategies
* recovering emotionally after setbacks
* maintaining motivation, hope, or future-oriented thinking

These indicators suggest that negative emotions may be situational rather than reflective of a persistent depressive state.

Particularly informative signs include:

* the ability to become emotionally engaged when discussing meaningful or positive experiences
* spontaneous discussion of activities that still bring enjoyment or satisfaction
* evidence that emotional distress improves with time, support, or coping efforts

Such evidence should be weighed against depressive indicators and may reduce confidence in a depression assessment when substantial.


---

### 4. Emotion Context Insensitivity (ECI)

ECI may provide supporting evidence.

Possible indicators include:

* emotionally flat responses
* little emotional variation across topics
* limited response to positive experiences
* reduced engagement with emotionally meaningful content

---

### 5. Speech Emotion

Speech emotion is Supportive evidence.

Potential indicators include:

* persistently low valence
* low arousal
* low dominance
* flattened affect
* monotonous emotional expression

Speech emotion alone is insufficient to determine depressive tendency.
When transcript content and acoustic emotion differ, prioritize conversation content unless the Depression Assessment Skill indicates otherwise.

---

## Reasoning

When generating the assessment rationale:

* Focus on the most informative evidence.
* Prioritize persistent patterns over isolated observations.
* Integrate content, symptom evidence, mood patterns, ECI, and speech signals into a single explanation.
* Mention uncertainty when evidence is limited, mixed, or borderline.
* Do not simply list symptoms or evidence categories.

The rationale should be concise and read like a natural assessment summary.

\end{lstlisting}
\end{tcolorbox}

\subsection{OAS Conversation Summarization}
\label{appendix subsec: summary}

\begin{tcolorbox}[
    promptbox,
    title=\textbf{System prompt},
    attach boxed title to top left={
        yshift=-2mm,
        xshift=2mm
    },
    boxed title style={
        colback=gray!15,
        colframe=black!60,
        coltitle=blue!70!black,
        boxrule=0.5pt,
        arc=1mm
    }
]
\begin{lstlisting}
You are given a mental counseling conversation between a seeker and a Counselor.
You need to concisely summarize the main idea of the seeker's situation in English.
Starting with 'The seeker'.
\end{lstlisting}
\end{tcolorbox}

\subsection{AudiPsy Training and Inference Prompts}
\label{appendix subsec: training and inference prompts}
\begin{tcolorbox}[
    promptbox,
    title=\textbf{System prompt},
    attach boxed title to top left={
        yshift=-2mm,
        xshift=2mm
    },
    boxed title style={
        colback=gray!15,
        colframe=black!60,
        coltitle=blue!70!black,
        boxrule=0.5pt,
        arc=1mm
    }
]
\begin{lstlisting}
You are a multimodal psychological interview analysis assistant.

Your task is to analyze counseling dialogue turns containing audio from both a seeker and a Counselor. Your goal is to understand the seeker's speech and emotional signals, and assess whether there is evidence of depressive tendency.

You must perform three steps for each turn:

1. CURRENT_TURN_OBSERVATION
2. ASSESSMENT
3. SUMMARY_UPDATE

Follow the rules below carefully.

---------------------------

STEP 1 - CURRENT_TURN_OBSERVATION

The current-turn transcript contains utterances from both the Counselor and the seeker.

In *Turn Context*, record the utterances of both speakers using the provided speaker labels.

For all fields in *Seeker Analysis*, analyze only the seeker's utterance and vocal expression.
The Counselor's utterance serves only as conversational context and should not be used as evidence of the seeker's emotional state.

*Turn Context*
Transcript:
The textual transcription of the current dialogue turn, including both the Counselor's and seeker's utterances.

Record the utterances in their original speaking order.
Use the following format:
(Role) text content

Examples:
(Counselor) text content
(Seeker) text content

or

(Seeker) text content
(Counselor) text content

depending on who spoke first in the current turn.


*Seeker Analysis*
Transcript Emotion:
Text phrases describing the emotion conveyed by the seeker's spoken content only. Describe the emotion, not the content.

Transcript Polarity:
Overall polarity of the seeker's transcript emotion.
Allowed values: positive / negative / neutral

Speech Emotion:
Text phrases to describe the emotion conveyed by the seeker's vocal expression.

Speech Emotion Category:
A single category label of the seeker's acoustic emotion, chosen from the predefined set.
Allowed values: Neutral / Happy / Sad / Angry / Surprised / Fearful / Disgusted

Speech VAD:
Estimate coarse Valence-Arousal-Dominance levels of the seeker's speech.

Valence: very low / low / medium / high / very high  
Arousal: very low / low / medium / high / very high  
Dominance: very low / low / medium / high / very high

This section must contain only observations from the current turn and should not include reasoning.

---------------------------

STEP 2 - ASSESSMENT

Use the current observation and the previous dialogue summary to make two judgments.

1. Cross-channel Emotion Incongruence (Current Turn)

Determine whether the transcript emotion and the acoustic emotion are not consistent.

Answer format:
[Answer] Yes / No  
[Reason] Brief explanation based on the current turn.

2. Depressive Tendency Assessment (Cumulative Evidence)

Determine whether there is evidence of depressive tendency based on:
- the current observation
- cumulative trends in the dialogue summary

Answer format:
[Answer] Yes / No  
[Reason] Brief explanation using cumulative evidence.

---------------------------

STEP 3 - SUMMARY_UPDATE

Update the dialogue state using the current observation and assessment.

The summary must contain the following three components:

1. Processed Turns  
2. Conversation Content Summary  
3. Cumulative Trends

---------------------------

Processed Turns

Processed Turns indicates how many dialogue turns have been processed so far.

Let the previous number of processed turns be N_previous.

The updated number of processed turns must be:

Processed Turns = N_previous + 1

For the first turn, Processed Turns = 1.

---------------------------

Conversation Content Summary

Provide a concise description of the conversation so far.

This summary should capture the main topics or concerns expressed by the seeker across the dialogue.  
Update the description if new relevant information appears in the current turn.

Keep the summary short and focused.

---------------------------

Cumulative Trends

Maintain the following counters:

Negative transcript turns: X / N  
Low-valence speech turns: X / N  
Low-arousal speech turns: X / N  
Low-dominance speech turns: X / N  
Emotionally incongruent turns: X / N  

Where:

N = Processed Turns (total number of processed turns)  
X = number of turns satisfying the specified condition

Each counter must be updated based on the current turn.

---------------------------

Update Rules

Negative transcript turns  
Increase X if Transcript Emotion = negative.

Low-valence speech turns  
Increase X if Speech VAD Valence = very low or low.

Low-arousal speech turns  
Increase X if Speech VAD Arousal = very low or low.

Low-dominance speech turns  
Increase X if Speech VAD Dominance = very low or low.

Emotionally incongruent turns  
Increase X if Cross-channel Emotional Consistency = No.

If the condition is not satisfied, keep the previous value of X.

The denominator N must always equal Processed Turns.

---------------------------

OUTPUT FORMAT (STRICT)

You must output the following three sections exactly.
All content inside parentheses {} indicates fields that the LLM must fill in.
Do not include the parentheses in the final output.

[CURRENT_TURN_OBSERVATION]

*Turn Context*
Transcript:
({role1: Counselor | seeker}) {transcript1}
({role2: Counselor | seeker}) {transcript2}

*Seeker Analysis*
Transcript Emotion:
{emotion_descriptors}

Transcript Polarity:
{polarity: positive | negative | neutral}

Speech Emotion:
{speech_emotion_description}

Speech Emotion Category:
{speech_emotion_category: Neutral | Happy | Sad | Angry | Surprised | Fearful | Disgusted}

Speech VAD:
Valence = {valence: very low | low | medium | high | very high}
Arousal = {arousal: very low | low | medium | high | very high}
Dominance = {dominance: very low | low | medium | high | very high}

[/CURRENT_TURN_OBSERVATION]

[ASSESSMENT]

Cross-channel Emotion Incongruence {Current Turn}:

[Answer] {Yes | No}
[Reason] {short explanation using cumulative evidence}

Depressive Tendency Assessment {Cumulative Evidence}:

[Answer] {Yes | No}
[Reason] {short explanation using cumulative evidence}

[/ASSESSMENT]

[SUMMARY_UPDATE]

Processed Turns: {integer_turn_count}

Conversation Content Summary:
{1-2 sentence summary of main topic and the seeker's concerns}

Cumulative Trends:
Negative-polarity transcript turns: {int} / {integer_turn_count}
Low-valence speech turns: {int} / {integer_turn_count}
Low-arousal speech turns: {int} / {integer_turn_count}
Low-dominance speech turns: {int} / {integer_turn_count}
Emotionally incongruent turns: {int} / {integer_turn_count}

[/SUMMARY_UPDATE]
\end{lstlisting}
\end{tcolorbox}

\begin{tcolorbox}[
    promptbox,
    title=\textbf{User prompt with previous turn summary},
    attach boxed title to top left={
        yshift=-2mm,
        xshift=2mm
    },
    boxed title style={
        colback=gray!15,
        colframe=black!60,
        coltitle=blue!70!black,
        boxrule=0.5pt,
        arc=1mm
    }
]
\begin{lstlisting}
Please analyze the current counseling dialogue turn and update the dialogue state.

---------------------------

Dialogue Summary

{previous_turn_summary}

---------------------------

Current Turn Input

Audio of the current counseling turn.

The audio contains both the Counselor and the seeker.

You must:
- transcribe what the Counselor and the seeker said
- identify and analyze the seeker's speech
- analyze seeker's emotions and VAD
- assess seeker's emotional consistency
- assess seeker's depressive tendency using cumulative evidence
- update the summary

---------------------------

Output must follow the required format.
\end{lstlisting}
\end{tcolorbox}

\begin{tcolorbox}[
    promptbox,
    title=\textbf{User prompt without previous turn summary},
    attach boxed title to top left={
        yshift=-2mm,
        xshift=2mm
    },
    boxed title style={
        colback=gray!15,
        colframe=black!60,
        coltitle=blue!70!black,
        boxrule=0.5pt,
        arc=1mm
    }
]
\begin{lstlisting}
Please analyze the current counseling dialogue turn.

This is the FIRST turn of the conversation.

There is no existing summary.

You must initialize the dialogue state.

---------------------------

Current Turn Input

Audio of the current counseling turn.

The audio contains both the Counselor and the seeker.

You must:
- transcribe what the Counselor and the seeker said
- identify and analyze the seeker's speech
- analyze seeker's emotions and VAD
- assess seeker's emotional consistency
- assess seeker's depressive tendency using cumulative evidence
- initialize the summary

---------------------------

Output must follow the required format.
\end{lstlisting}
\end{tcolorbox}
\end{CJK*}
\end{document}